\documentclass[10pt,twocolumn,letterpaper]{article}
\usepackage[T1]{fontenc}
\usepackage[pagenumbers]{cvpr}
\usepackage{graphicx}
\usepackage{amsmath,amssymb,booktabs,xcolor,array,colortbl,tabularx,multirow}
\definecolor{cvprblue}{rgb}{0.21,0.49,0.74}
\usepackage{listings,tcolorbox}
\definecolor{promptblue}{HTML}{EEF5FC}
\definecolor{promptgreen}{HTML}{EFF6EF}
\newcommand{\origans}{\textit{Original + answer}}
\newcommand{\evidencecheck}{\textcolor{green!45!black}{\ensuremath{\checkmark}}}
\newcolumntype{Y}{>{\centering\arraybackslash}X}
\usepackage[pagebackref,breaklinks,colorlinks,allcolors=cvprblue]{hyperref}
\DeclareRobustCommand{\docref}[2]{\ref{#1}}
\def\confName{CVPR}
\def\confYear{2027}
\newcommand{\sg}{\operatorname{sg}}
\newcommand{\clip}{\operatorname{clip}}
\newcommand{\draftnote}[1]{}
\title{Seeing the Right Context: Grounded On-Policy Distillation for Remote Sensing}
\author{%
  Yichang Luo\textsuperscript{1,2$*$},
  Hourun Zhu\textsuperscript{3$*$},
  Shixiong Liu\textsuperscript{1$\dagger$},
  Kan Wei\textsuperscript{1,2},
  Shi Liang\textsuperscript{4},
  Lei Wang\textsuperscript{1,2}\\[4pt]
  \textsuperscript{1}Aerospace Information Research Institute, Chinese Academy of Sciences\\
  \textsuperscript{2}University of Chinese Academy of Sciences\quad
  \textsuperscript{3}Beijing Institute of Technology\\
  \textsuperscript{4}Huazhong University of Science and Technology
}
\begin{document}
\maketitle
\begingroup
\renewcommand\thefootnote{\fnsymbol{footnote}}
\footnotetext[1]{Equal contribution.}
\footnotetext[2]{Corresponding author.}
\endgroup

\setlength{\textfloatsep}{10pt plus 2pt minus 2pt}
\setlength{\floatsep}{8pt plus 2pt minus 2pt}
\setlength{\intextsep}{8pt plus 2pt minus 2pt}

\raggedbottom

\begin{abstract}
Remote sensing scene understanding requires multimodal large language models (MLLMs) to interpret queried targets in relation to nearby objects and surrounding structures, which serve as spatial references or define the scope of an answer. However, standard answers and bounding boxes often leave these contextual relationships implicit. Thus, we construct GeoContext-5K to make them explicit in 5,394 training pairs for visual question answering (VQA), with spatial cues, visual facts, and supporting regions. GeOPD uses these annotations for on-policy distillation, guiding a frozen teacher with region marks on the full image and a visual fact while preserving surrounding spatial references. It prioritizes teacher corrections that agree with the shift in the student’s next token distribution when evidence is restored. A complementary objective weights the contrast between the student’s scores on the two views by how much masking lowers the teacher’s log probability for the same sampled token. Experiments on five remote sensing VQA benchmarks show that GeOPD outperforms the evaluated open-source MLLMs and matched post-training baselines, improving over the 4B and 9B base models by 5.81 and 2.97 percentage points on average.
The dataset and code will be publicly released at
\href{https://github.com/YeeCharm/GeOPD}{\textcolor{cvprblue}{\textit{https://github.com/YeeCharm/GeOPD}}}.
\end{abstract}

\section{Introduction}
\begin{figure*}[tbp]
\centering
\includegraphics[width=\textwidth]{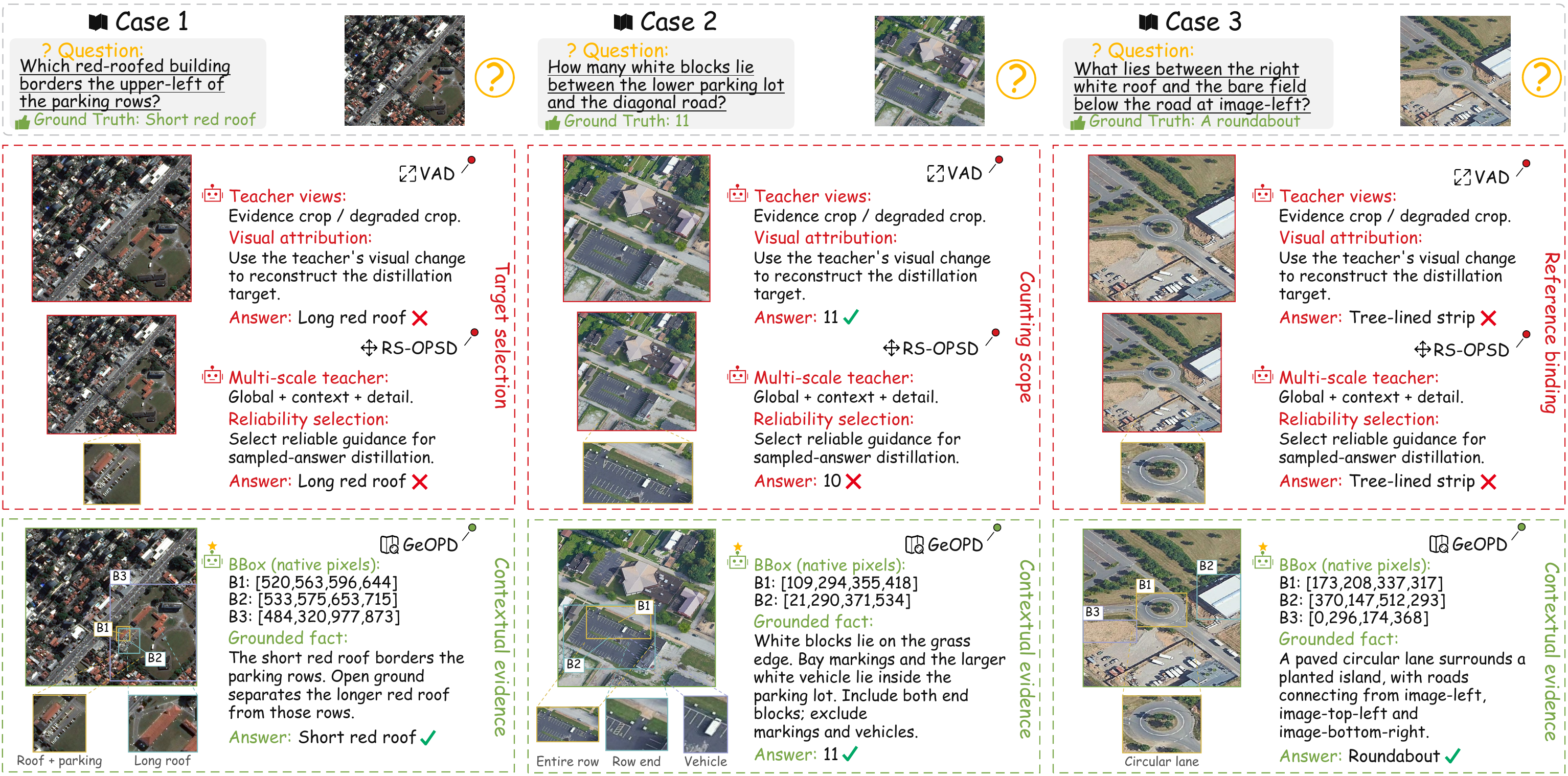}
\captionsetup{skip=4pt}
\caption{\textbf{Scene context determines what counts as evidence.} Three examples compare VAD, RS-OPSD, and GeOPD on target selection, counting within a specified region, and reasoning from landmarks. The annotations and crops identify evidence for each question within its scene.}
\label{fig:motivation}
\end{figure*}

Remote sensing scene understanding involves recognizing objects, interpreting spatial relations, and characterizing larger regions~\cite{lobry2020rsvqa,wang2024earthvqa}. In wide aerial scenes where similar objects recur~\cite{wang2025xlrs}, MLLMs must interpret the scene in relation to the question, using nearby landmarks to identify the intended instance and surrounding structures to define the scope of spatially constrained tasks (\textit{e.g.}, the parking lot and diagonal road in Figure~\ref{fig:motivation}
delimit the area in which the white blocks are counted). The query thus determines which scene elements must be interpreted jointly.

Existing MLLMs often fail in such scenes when fine details and contextual cues are not effectively grounded. Prior work improves detail access via cropping and multi-view inspection~\cite{liu2026zoomearth,yuan2026vision,jiang2026rs}, yet tight crops often discard the external landmarks or boundaries needed to interpret a target, whereas full images leave relevant context unguided. Moreover, standard answer supervision and bounding boxes~\cite{hudson2019gqa,li2024vrsbench} rarely distinguish the target, the spatial reference, and the visual relation linking them.

Thus, we construct GeoContext-5K, a dataset of 5,394 training QA pairs (2,658 multiple-choice and 2,736 short-answer questions) across six task families, including region-bounded counting, spatial relations, and attribute comparisons. Each record provides a spatial cue indicating where to look, a visual fact describing what is visible, and grounding boxes distinguishing target evidence from spatial references. Drawing on VRSBench~\cite{li2024vrsbench}, LRS-GRO~\cite{liu2026zoomearth}, and RS5M~\cite{zhang2024rs5m}, these structured annotations let us anchor the teacher on the full scene while selectively occluding target evidence, preserving the surrounding landmarks and spatial boundaries required to resolve the query.

Building on this foundation, we further propose GeOPD, an on-policy distillation (OPD) framework that transfers contextual reasoning from a privileged teacher to the student's next-token predictions~\cite{agarwal2024policy}. Unlike crop-based or multi-view OPD strategies~\cite{jiang2026rs,yuan2026vision,zhang2026vad}, our frozen teacher receives a visual fact and the full scene with the target and reference regions marked, preserving the global layout. To assess the visual support for teacher corrections, we measure the shift in the student's next token distribution from the target-masked to the original view, holding the question and response prefix fixed. Corrections that align more closely with this shift receive greater weight in distillation.

To ensure the student actively relies on the target evidence rather than language priors, we introduce a visual contrast objective that compares its scores on the original and masked views. The frozen teacher scores the same sampled token on both views using only the corresponding image, question, and shared prefix. A larger drop in the teacher’s log probability under masking gives more weight to reinforcing the student’s existing score advantage on the original view. We combine this contrast loss with directionally weighted distillation, updating the student solely through the original image branch.

Our main contributions are summarized as follows:
\begin{itemize}
     \item We construct GeoContext-5K, a remote sensing dataset that explicitly disentangles target evidence from spatial references via spatial cues, atomic visual facts, and role-specific boxes across 5,394 verified pairs.
     \item We propose GeOPD to distill full-scene contextual evidence by weighting distribution matching via student visual alignment and calibrating cross-view contrast via teacher masking sensitivity.
     \item Across five remote sensing VQA benchmarks, GeOPD leads all post-training baselines at 4B (63.80\%) and 9B (64.98\%), with our 4B student surpassing the 9B base model by 1.79 points.

\end{itemize}

\section{Related Work}

\subsection{Context in Remote Sensing Understanding}
The surrounding layout helps resolve ambiguity in remote sensing imagery (\textit{e.g.}, identifying a runway from the taxiways that connect it to an apron). Existing models span scene description, VQA, and visual grounding~\cite{kuckreja2024geochat,pang2025vhm,li2024vrsbench,xin2026segearth,pmlr-v267-shabbir25a}. ZoomEarth~\cite{liu2026zoomearth} selects regions from a scene overview for closer inspection, GeoViS~\cite{zhang2026geovis} refines localization via hierarchical search, and recent work grounds intermediate reasoning in visible regions~\cite{liu2026towards}. While localization and zooming expose local details, the roles of surrounding landmarks or boundaries remain implicit, motivating supervision that specifies how target evidence and spatial context jointly support an answer. 


\subsection{Privileged On-Policy Distillation}
On-policy distillation supervises the student along its own sampled trajectories~\cite{gu2024minillm,agarwal2024policy}, with privileged variants augmenting teacher inputs during training. Vision-OPD~\cite{yuan2026vision} supplies an evidence-centered crop, RP-OPSD~\cite{zhu2026rp} feeds native resolution images to the teacher and RS-OPSD~\cite{jiang2026rs} combines global, contextual, and detailed views. Textual guidance offers another form of privilege through spatial hints, cues recoverable from the image, or observations drawn from student trajectories~\cite{sirko2026opd,tian2026vicur,wei2026policy}. However, teacher corrections may reflect cropping artifacts~\cite{hu2026evidence} or ungrounded text priors, leaving open which corrections reflect evidence visible to the student.

\begin{figure*}[tbp]
\centering
\includegraphics[width=\textwidth]{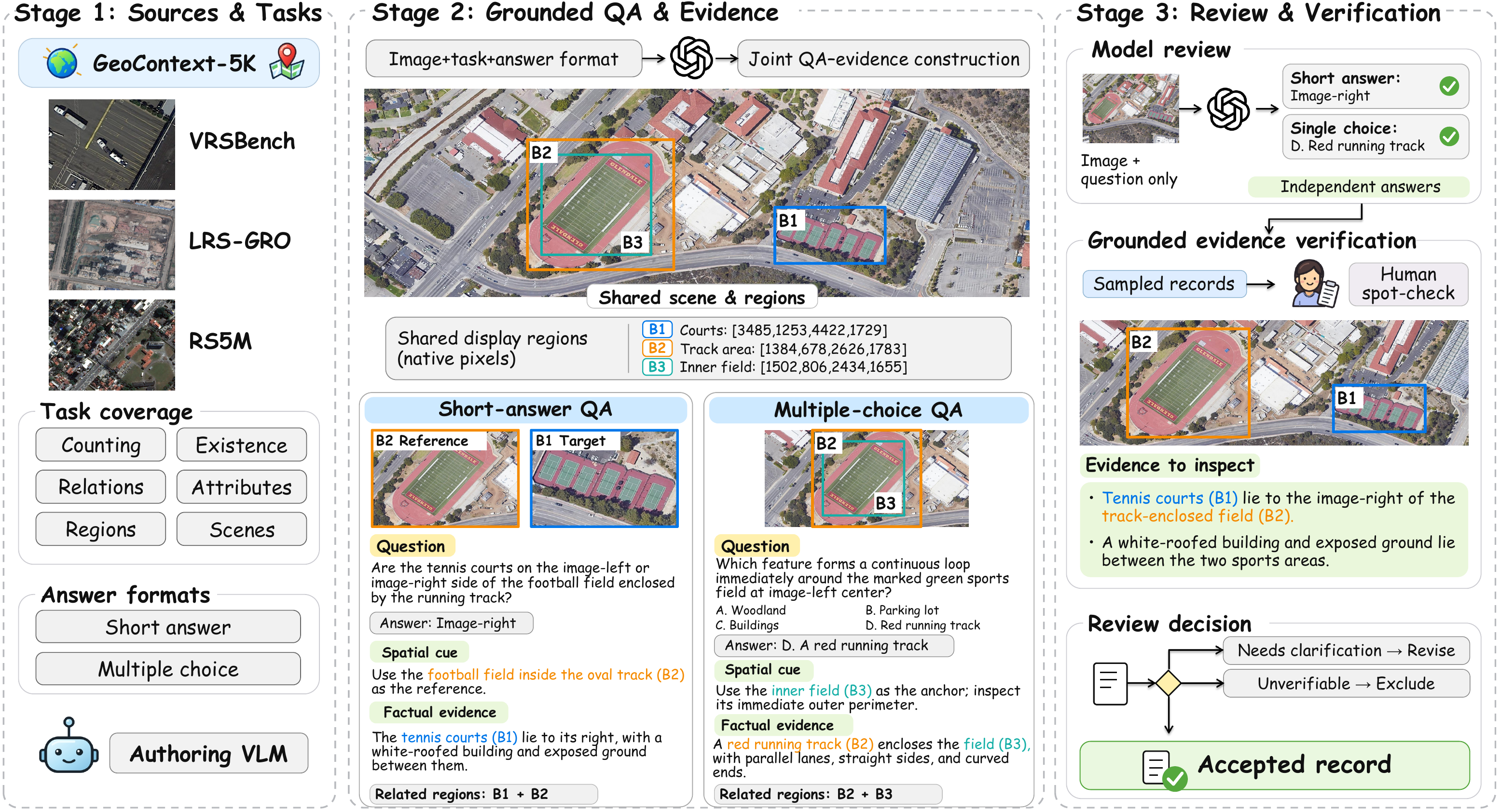}
\captionsetup{skip=4pt}
\caption{\textbf{Overview of GeoContext-5K construction.} The annotation pipeline comprises three stages covering source and task selection, joint construction of QA pairs and supporting evidence, and record verification. An independent model reviews the resulting records, and human reviewers inspect a subset.}
\label{fig:data}
\end{figure*}

\subsection{Distillation with Visual Evidence}
Counterfactual visual perturbations offer a principled mechanism to probe how model predictions rely on specific visual evidence~\cite{niu2021counterfactual,leng2024mitigating}, serving as an effective vehicle to refine teacher supervision during distillation~\cite{qian2026med,wang2026cv,feng2026vc,xue2026distill}. Along this line, VA-OPD~\cite{liu2026visual} reweights distillation by detail sensitivity, VAD~\cite{zhang2026vad} reconstructs targets via visual attribution, EAD~\cite{hu2026evidence} aligns teacher corrections with student shifts under evidence restoration, and S-OPD~\cite{liu2026better} calibrates cross-view student contrast via teacher masking drops. Despite their effectiveness, these approaches typically treat visual evidence monolithically, overlooking the interplay between target instances and surrounding spatial context. In this work, we leverage question-specific scene annotations to anchor a full-scene teacher while explicitly decoupling target object masks from spatial references, providing fine-grained, disentangled supervision for both objectives.

\section{GeoContext-5K Dataset}
\label{sec:data}

Existing remote sensing VQA and grounding datasets~\cite{li2024vrsbench,liu2026zoomearth} rarely make explicit how surrounding landmarks and boundaries support individual answers. We construct GeoContext-5K via the three-stage pipeline in Figure~\ref{fig:data} to link queries and answers to their target evidence and spatial references in the full scene (see Appendix~\docref{app:data}{A} for dataset comparisons and review criteria).

\noindent\textbf{Stage 1. Sources and Tasks.} GeoContext-5K draws on imagery and available annotations from VRSBench~\cite{li2024vrsbench}, LRS-GRO~\cite{liu2026zoomearth}, and RS5M~\cite{zhang2024rs5m}. The source image is the unit of sampling, and each candidate retains the full scene together with any available descriptions and region annotations. Task selection reflects what is visible at the scale of the intended question, from a target and its spatial references to the layout of a wider region. The six task families cover counting, existence and negation, spatial relations, attributes and comparisons, regional extent and boundaries, and scenes and land cover.

\noindent\textbf{Stage 2. Grounded QA and Evidence.} We use GPT-6.1 Sol~\cite{openai2026gpt61sol} to construct QA pairs, with the procedure depending on whether a suitable question is already available. We retain or revise source questions that support a definite answer from the image, or draft new QA pairs otherwise. To eliminate referential ambiguity, spatial relation queries explicitly specify the interacting entities, whereas counting queries strictly define both the target category unit and the spatial search scope, with all instances formatted as concise short-answer or multiple-choice items. For fine-grained supervision, each data record couples the question--answer pair with structured multi-level annotations: (i) a textual spatial cue that designates the target entity, reference anchor, or search region; (ii) an atomic visual fact articulating the underlying observational evidence; and (iii) precise bounding boxes that separately localize the target evidence and its corresponding spatial references (\textit{e.g.}, the running track in Figure~\ref{fig:data} locates the reference field for the tennis court question and is itself the target feature to identify in the question about the field's perimeter).

\begin{figure*}[tbp]
\centering
\includegraphics[width=\textwidth]{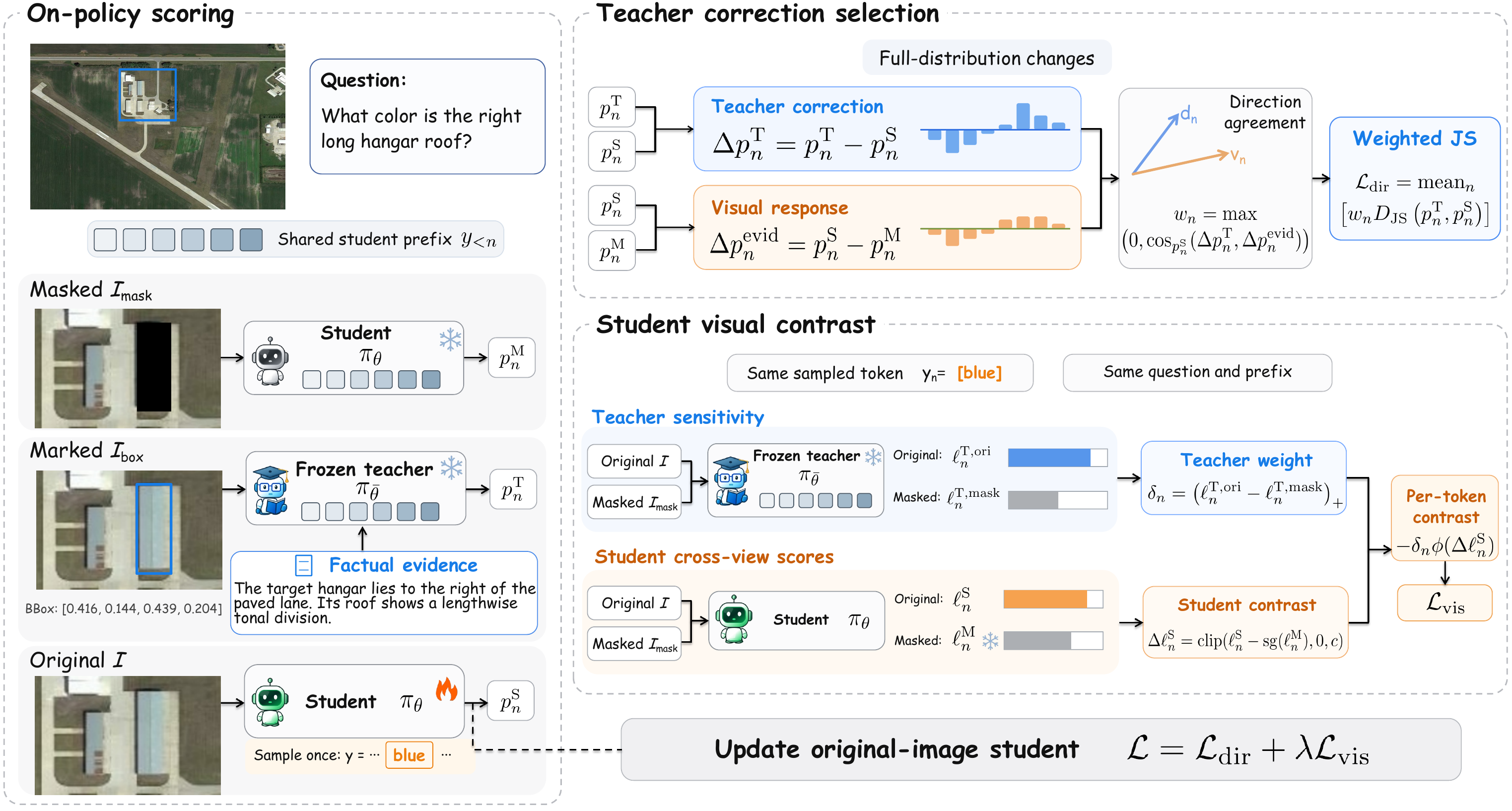}
\captionsetup{skip=4pt}
\caption{\textbf{Overview of GeOPD.} \textbf{Left:} The student samples a response on the original image. At each prefix, it predicts from the original and masked views, and the teacher from the marked view with a visual fact. \textbf{Right:} Agreement between teacher corrections and the student's response to restored evidence weights distillation. The frozen teacher's score drop under masking scales contrast for the sampled token. Both losses backpropagate through the original image branch. For the single sample shown, we omit the response index $i$.}
\label{fig:method}
\end{figure*}

\noindent\textbf{Stage 3. Review and Verification.} To verify that the answers and annotations are supported by the scene, an independent model first answers from the unmarked image and the question alone. It then compares its response with the reference answer and checks that the referent and scope are clear, the stated facts are visible, and the boxes cover the described evidence. Following human spot-checks, we revise ambiguous records and discard unverifiable ones. The final GeoContext-5K dataset contains 5,394 verified training QA pairs, comprising 2,658 multiple-choice questions and 2,736 short-answer questions, with dual-tier perceptual hash and feature similarity checks screening for potential visual overlaps against downstream evaluation benchmarks.

\section{GeOPD}
\label{sec:method}

Building on GeoContext-5K, GeOPD uses the evidence annotated for each question to guide the teacher within the full scene and to isolate target evidence from spatial references for counterfactual masking. These annotations support two training objectives in Figure~\ref{fig:method}. Directional distillation upweights teacher corrections aligned with the student's response to restoring target evidence, while visual contrast uses teacher masking sensitivity to reinforce the student's original-view score.

\subsection{Contextual Evidence for Supervision}
To guide the teacher while preserving surrounding context, we mark supporting regions on the full image. In Figure~\ref{fig:method}, the box highlights the queried roof, and the visual fact locates the hangar relative to the paved lane. Combining the marked view $I_{\mathrm{box}}$ of image $I$ with question $q$ and visual fact $e_{\mathrm{fact}}$ gives the privileged context $c^T=(I_{\mathrm{box}},q,e_{\mathrm{fact}})$. While $I_{\mathrm{box}}$ marks both target and reference regions to anchor the teacher, we construct $I_{\mathrm{mask}}$ by filling only target evidence pixels in $I$ with the scene's mean RGB color to avoid black-box artifacts. Crucially, surrounding landmarks, search boundaries, and any reference regions overlapping or nested within the target box (\textit{e.g.}, the inner field within the running track in Figure~\ref{fig:data}) remain unoccluded, isolating the visibility of the queried target.
For each rollout, the student $\pi_\theta$ samples a response $y$ from the original image and question. Scoring along that response keeps the generated text fixed across views. At each position $n$, the original student, privileged teacher, and masked student produce three distributions, all conditioned on the same prefix $y_{<n}$:

\begin{equation}
\begin{aligned}
 p_n^{\mathrm{S}}&=\pi_\theta(\cdot\mid I,q,y_{<n}),\\
 p_n^{\mathrm{T}}&=\pi_{\bar\theta}(\cdot\mid c^T,y_{<n}),\\
 p_n^{\mathrm{M}}&=\pi_\theta(\cdot\mid I_{\mathrm{mask}},q,y_{<n}).
\end{aligned}
 \label{eq:distributions}
\end{equation}

Initialized from the same checkpoint as the student, the teacher $\pi_{\bar\theta}$ remains frozen throughout training. Both student passes use the current parameters $\theta$, with gradients stopped through the masked pass.

\subsection{Selecting Visually Aligned Corrections}
At each response position, the privileged teacher's correction $\Delta p_n^{\mathrm{T}}=p_n^{\mathrm{T}}-p_n^{\mathrm{S}}$ may reflect visual evidence, the textual fact $e_{\mathrm{fact}}$, or parameter drift. While evidence-aligned distillation~\cite{hu2026evidence} compares teacher shifts against generic full-evidence removals, remote sensing requires verifying whether a correction is grounded in the target instance under fixed spatial references. We therefore compare $\Delta p_n^{\mathrm{T}}$ against the student's probability shift under target-only restoration with all reference anchors preserved, $\Delta p_n^{\mathrm{evid}}=p_n^{\mathrm{S}}-p_n^{\mathrm{M}}$, sharing the same parameters, question, and prefix.

To account for the local curvature of the probability simplex under Jensen--Shannon (JS) divergence, where an identical probability shift carries greater sensitivity at lower baseline probabilities, we evaluate the inverse-probability weighted inner product across candidate tokens $k\in\mathcal K_n$:
\begin{equation}
 \big\langle \Delta p_n^{\mathrm{T}},\Delta p_n^{\mathrm{evid}}\big\rangle_{p_n^{\mathrm{S}}}
 = \sum_{k\in\mathcal K_n}
 \frac{\Delta p_n^{\mathrm{T}}(k)\,\Delta p_n^{\mathrm{evid}}(k)}
      {\max\!\left(p_n^{\mathrm{S}}(k),\epsilon\right)},
 \label{eq:directions}
\end{equation}
where $\epsilon>0$ bounds the denominator and $\|u\|_{p_n^{\mathrm{S}}}=\sqrt{\langle u,u\rangle_{p_n^{\mathrm{S}}}}$ denotes the induced local norm. We take the positive part of the normalized alignment to obtain the distillation weight at each position:
\begin{equation}
 w_n=\max\!\left(
 0,\frac{\big\langle \Delta p_n^{\mathrm{T}},\Delta p_n^{\mathrm{evid}}\big\rangle_{p_n^{\mathrm{S}}}}
        {\|\Delta p_n^{\mathrm{T}}\|_{p_n^{\mathrm{S}}}\,\|\Delta p_n^{\mathrm{evid}}\|_{p_n^{\mathrm{S}}}}
 \right).
 \label{eq:weight}
\end{equation}

Greater agreement across candidate tokens gives a larger $w_n$, whereas opposing updates or negligible variations on visually invariant syntactic tokens (where the product of local norms falls below $10^{-8}$) receive zero weight. We weight the JS divergence between the privileged teacher and the original student by $w_n$ and average over the valid positions $\mathcal N(y)$ of each response:
\begin{equation}
 \mathcal L_{\mathrm{dir}}=\mathbb E_y\!\left[
 \frac{1}{|\mathcal N(y)|}\sum_{n\in\mathcal N(y)}
 \sg(w_n)D_{\mathrm{JS}}\!\left(\sg(p_n^{\mathrm{T}}),p_n^{\mathrm{S}}\right)
 \right].
 \label{eq:distill}
\end{equation}

During the update, $\sg$ stops gradients through the teacher distribution and the selection weight, and the masked pass serves as a fixed reference. Averaging over valid positions lets the weights control both the token-level allocation of supervision and the overall strength of distillation.

\subsection{Student Visual Contrast}
\label{sec:contrast}

A student may match the teacher on the original image yet respond weakly to target removal. Building on teacher-calibrated policy contrast~\cite{liu2026better}, we contrast the student's scores on the original and target-masked views, upweighting positions where target removal under preserved spatial context causes a larger drop in the teacher's score.

At position $n$ of response $i$, the frozen teacher scores the sampled token $y_{in}$ on $I_i$ and $I_{i,\mathrm{mask}}$ without privileged marks or facts, receiving only the corresponding image, question $q_i$, and shared prefix $y_{i,<n}$. Given the corresponding log probabilities $\ell_{in}^{\mathrm{T,ori}}$ and $\ell_{in}^{\mathrm{T,mask}}$, we take the positive part of their difference as the sensitivity weight:
\begin{equation}
 \delta_{in}=\sg\!\left[\left(
 \ell_{in}^{\mathrm{T,ori}}-\ell_{in}^{\mathrm{T,mask}}
 \right)_+\right].
 \label{eq:visualgate}
\end{equation}

For the sampled color token in Figure~\ref{fig:method}, $\delta_{in}$ is positive if hiding the roof lowers the teacher's score for \emph{blue}. This weight acts on the score comparison for the sampled token, while $w_n$ controls matching of the next token distribution at the same position.

To measure the student's visual evidence gain at the same position, we evaluate $\ell_{in}^{\mathrm{S}}=\log p_{in}^{\mathrm{S}}(y_{in})$ and $\ell_{in}^{\mathrm{M}}=\log p_{in}^{\mathrm{M}}(y_{in})$. Unlike two-sided contrast objectives that can inadvertently depress original-view scores when $\ell_{in}^{\mathrm{S}}<\ell_{in}^{\mathrm{M}}$, we define the one-sided log-likelihood gain $\Delta\ell_{in}^{\mathrm{S}}=\clip(\ell_{in}^{\mathrm{S}}-\sg(\ell_{in}^{\mathrm{M}}),0,c)$. Using the bounded gain shaping function $\phi(\Delta\ell_{in}^{\mathrm{S}})=\min\{K,\exp(-\Delta\ell_{in}^{\mathrm{S}})-1+\Delta\ell_{in}^{\mathrm{S}}\}$, where $c$ and $K$ bound the contribution of large margins while keeping gradients zero-grounded when $\ell_{in}^{\mathrm{S}}\leq\ell_{in}^{\mathrm{M}}$, we formulate the contrast loss over $R$ responses in a batch as:
\begin{equation}
 \mathcal L_{\mathrm{vis}}=-\frac{1}{R}\sum_{i=1}^{R}
 \frac{\sum_{n\in\mathcal A_i}\delta_{in}\,\phi(\Delta\ell_{in}^{\mathrm{S}})}
      {\max(1,|\mathcal A_i|)}.
 \label{eq:contrast}
\end{equation}

Here $\mathcal A_i$ contains valid positions with $\delta_{in}>0$, excluding the end token. We normalize by the number of active positions in each response, with zero contribution from responses without active positions. With the teacher weight and masked score detached, the gradient with respect to $\ell_{in}^{\mathrm{S}}$ is $-\delta_{in}(1-\exp(-\Delta\ell_{in}^{\mathrm{S}}))\leq 0$ when $\ell_{in}^{\mathrm{S}}>\ell_{in}^{\mathrm{M}}$ and zero otherwise. Thus, $\mathcal L_{\mathrm{dir}}$ first steers the full distribution toward evidence-backed tokens, while $\mathcal L_{\mathrm{vis}}$ conditionally sharpens confidence only on validated visual margins, preventing ungrounded reinforcement.

\begin{table*}[tbp]
\centering\small
\renewcommand{\arraystretch}{0.95}
\captionsetup{skip=4pt}
\caption{\textbf{Comparison with open-source models.} Scores are percentages. Average weights the five benchmarks equally. Bold and underlining mark the best and second best scores in each column.}
\label{tab:main}
\setlength{\tabcolsep}{3pt}
\begin{tabular*}{\textwidth}{@{\extracolsep{\fill}}lcccccccc@{}}\toprule
Model & Pub. & Params. & CHOICE & GEOBench-VLM & OmniEarth & HRVQA & LHRS-Bench & Average\\\midrule
\multicolumn{9}{c}{\textit{Open-Source General Models}}\\
\midrule
InternVL2.5~\cite{chen2024expanding} & arXiv 2024 & 8B & 71.79 & 45.66 & 48.03 & 28.55 & 71.16 & 53.04\\
InternVL3.5~\cite{wang2025internvl3} & arXiv 2025 & 8B & 72.90 & 48.99 & 47.58 & 37.10 & 72.03 & 55.72\\
Qwen2.5-VL-Instruct~\cite{bai2025qwen25vltechnicalreport} & arXiv 2025 & 7B & 75.09 & 48.43 & 48.13 & 41.10 & 69.57 & 56.46\\
MiniCPM-V-4.5~\cite{Yu_2026_CVPR} & CVPR 2026 & 8B & 76.56 & 50.36 & 50.24 & 37.20 & 73.33 & 57.54\\
MiMo-VL-RL~\cite{yue2025mimo} & arXiv 2025 & 7B & 76.24 & 46.25 & 46.75 & 44.80 & 70.58 & 56.92\\
GLM-4.6V-Flash~\cite{hong2025glm} & arXiv 2025 & 9B & 76.85 & 50.89 & 55.83 & 42.80 & 73.62 & 60.00\\
ZwZ~\cite{pmlr-v306-wei26j} & ICML 2026 & 8B & 76.29 & 50.26 & 49.73 & 34.90 & 75.80 & 57.40\\
Qwen3-VL~\cite{bai2025qwen3} & arXiv 2025 & 8B & 78.01 & 49.67 & 52.67 & 42.60 & 74.35 & 59.46\\
Qwen3.5~\cite{qwen3.5} & arXiv 2026 & 4B & 78.52 & 48.89 & 48.47 & 38.55 & 75.51 & 57.99\\
Qwen3.5~\cite{qwen3.5} & arXiv 2026 & 9B & \underline{81.54} & 51.39 & 55.08 & 42.75 & \underline{79.28} & 62.01\\
\midrule
\multicolumn{9}{c}{\textit{Open-Source Remote Sensing Models}}\\
\midrule
GeoChat~\cite{kuckreja2024geochat} & CVPR 2024 & 7B & 57.39 & 42.29 & 30.91 & 27.25 & 51.88 & 41.94\\
SkySenseGPT~\cite{luo2024skysensegpt} & arXiv 2024 & 7B & 61.70 & 42.57 & 30.53 & 31.95 & 51.59 & 43.67\\
EarthMind~\cite{shu2025earthmind} & arXiv 2025 & 4B & 73.03 & 44.69 & 39.89 & 19.60 & 70.58 & 49.56\\
VHM~\cite{pang2025vhm} & AAAI 2025 & 7B & 67.32 & 46.40 & 33.09 & 27.75 & 53.19 & 45.55\\
EarthDial-RGB~\cite{soni2025earthdial} & CVPR 2025 & 4B & 67.13 & 50.02 & 36.07 & 35.15 & 63.77 & 50.43\\
RemoteReasoner~\cite{yao2026remotereasoner} & AAAI 2026 & 7B & 74.26 & 48.18 & 47.16 & 37.15 & 67.10 & 54.77\\
TerraScope~\cite{Shu_2026_CVPR} & CVPR 2026 & 8B & 47.01 & 35.07 & 25.42 & 26.55 & 47.25 & 36.26\\\midrule
GeOPD & -- & 4B & 81.39 & \textbf{52.85} & \underline{55.84} & \underline{49.20} & \textbf{79.71} & \underline{63.80}\\
GeOPD & -- & 9B & \textbf{83.59} & \underline{52.41} & \textbf{58.07} & \textbf{52.40} & 78.41 & \textbf{64.98}\\\bottomrule
\end{tabular*}
\end{table*}

Directional distillation supplies the signal for matching the privileged teacher. We combine both losses, with $\lambda$ controlling the contribution of visual contrast:
\begin{equation}
 \mathcal L=\mathcal L_{\mathrm{dir}}+\lambda\mathcal L_{\mathrm{vis}}.
 \label{eq:total}
\end{equation}

Both terms update the student through the original image branch. At inference, the student answers from the original image and question alone.

\section{Experiments}
\label{sec:experiments}

\subsection{Experimental Setup}
\noindent\textbf{Training setup.} We train Qwen3.5-4B/9B~\cite{qwen3.5} with GeOPD using 5,394 questions from GeoContext-5K. Directional distillation uses Jensen--Shannon (JS) divergence with a mixture coefficient of 0.5. We compute it over the 100 tokens ranked highest by the student on the original image and a single bucket for the remaining probability mass. The teacher stays frozen at the student's initial checkpoint, and the visual contrast coefficient is $\lambda=0.02$. The 4B OPD runs use one epoch, batches of 96 questions, four responses per question, and a learning rate of $2\times10^{-6}$. We train GeOPD for 56 optimization steps at both backbone sizes.

\noindent\textbf{Baselines.} Table~\ref{tab:main} compares GeOPD with open-source general and RS-specific MLLMs. For each backbone, Table~\ref{tab:strategies} evaluates SFT~\cite{liu2023visual}, GRPO~\cite{shao2024deepseekmath}, unprivileged self-distillation OPSD~\cite{agarwal2024policy}, and privileged or evidence-guided distillation methods including Vision-OPD~\cite{yuan2026vision}, RP-OPSD~\cite{zhu2026rp}, RS-OPSD~\cite{jiang2026rs}, VAD~\cite{zhang2026vad}, EAD~\cite{hu2026evidence}, and S-OPD~\cite{liu2026better}. All post-training methods share the initial checkpoint, training questions, budget, and corresponding GeoContext-5K annotations.

\noindent\textbf{Benchmarks.} Our evaluation spans object perception, spatial reasoning, and scene recognition across four multiple-choice benchmarks: CHOICE~\cite{an2026choice}, the single-image subset of GEOBench-VLM~\cite{danish2025geobench}, OmniEarth~\cite{fu2026omniearth}, LHRS-Bench~\cite{muhtar2024lhrs}, and one open-ended generation benchmark, HRVQA~\cite{li2024hrvqa}, on high-resolution aerial images. Appendix~\docref{app:evaluation}{C} details the evaluation protocols, image deduplication checks, and training efficiency.

\subsection{Main Results}
\label{sec:results}

\begin{table*}[tbp]
\centering\small
\renewcommand{\arraystretch}{0.97}
\captionsetup{skip=4pt}
\caption{\textbf{Comparison of post-training methods.} Trained methods share data and question budgets within each backbone group. GEO, Omni, and LHRS denote GEOBench-VLM, OmniEarth, and LHRS-Bench. Scores are percentages. Avg. weights the five benchmarks equally. Bold and underlining mark the best and second best scores within each group.}
\label{tab:strategies}
\setlength{\tabcolsep}{3pt}
\begin{tabular*}{\textwidth}{@{\extracolsep{\fill}}l*{6}{c}@{\hspace{8pt}}*{6}{c}@{}}\toprule
\multirow{2}{*}{Method} &
\multicolumn{6}{c}{\textit{Qwen3.5-4B Post-Training Strategies}} &
\multicolumn{6}{c@{}}{\textit{Qwen3.5-9B Post-Training Strategies}}\\
\cmidrule(lr){2-7}\cmidrule(lr){8-13}
& CHOICE & GEO & Omni & HRVQA & LHRS & Avg. & CHOICE & GEO & Omni & HRVQA & LHRS & Avg.\\\midrule
Base model & 78.52 & 48.89 & 48.47 & 38.55 & 75.51 & 57.99 & 81.54 & 51.39 & 55.08 & 42.75 & \underline{79.28} & 62.01\\
SFT & \underline{81.15} & 50.36 & \underline{53.85} & 46.95 & 78.26 & 62.12 & 82.92 & \textbf{52.73} & \underline{57.89} & 48.05 & 78.41 & \underline{64.00}\\
OPSD~\cite{agarwal2024policy} & 79.02 & 49.61 & 50.86 & 38.95 & 77.54 & 59.19 & 81.48 & 50.83 & 55.14 & 45.95 & 78.41 & 62.36\\
GRPO~\cite{shao2024deepseekmath} & 80.55 & 47.49 & 52.47 & 46.30 & 76.81 & 60.72 & 81.30 & 48.99 & 54.40 & \underline{51.65} & 77.54 & 62.78\\
Vision-OPD~\cite{yuan2026vision} & 79.86 & 49.86 & 51.54 & 39.80 & 78.26 & 59.86 & 81.86 & 50.11 & 54.64 & 44.75 & 78.99 & 62.07\\
RP-OPSD~\cite{zhu2026rp} & 80.20 & 47.96 & 53.76 & 42.40 & 79.28 & 60.72 & \underline{82.97} & 51.79 & 56.79 & 48.00 & 78.84 & 63.68\\
RS-OPSD~\cite{jiang2026rs} & 80.25 & 50.05 & 53.09 & \underline{48.40} & 79.28 & \underline{62.21} & 82.42 & 50.70 & 56.47 & 46.10 & 78.12 & 62.76\\
VAD~\cite{zhang2026vad} & 80.26 & \underline{50.58} & 53.39 & 43.05 & \textbf{80.29} & 61.51 & 81.91 & 50.26 & 56.76 & 47.00 & 78.99 & 62.98\\
EAD~\cite{hu2026evidence} & 79.93 & 49.58 & 51.43 & 39.15 & 78.12 & 59.64 & 81.88 & 50.45 & 54.80 & 45.15 & 78.70 & 62.19\\
S-OPD~\cite{liu2026better} & 79.80 & 49.80 & 51.97 & 40.10 & 78.55 & 60.04 & 81.75 & 51.14 & 55.51 & 45.25 & \textbf{79.42} & 62.61\\
GeOPD & \textbf{81.39} & \textbf{52.85} & \textbf{55.84} & \textbf{49.20} & \underline{79.71} & \textbf{63.80} & \textbf{83.59} & \underline{52.41} & \textbf{58.07} & \textbf{52.40} & 78.41 & \textbf{64.98}\\
\bottomrule
\end{tabular*}
\end{table*}

\noindent\textbf{Comparison with SOTA MLLMs.} GeOPD-9B and GeOPD-4B rank first and second by mean score among the open-weight models in Table~\ref{tab:main}. The 4B student improves over its base model by 5.81 percentage points on average and exceeds the original 9B model by 1.79 points. At both scales, the largest gains over the base model occur on open-ended HRVQA.

\noindent\textbf{Comparison with post-training methods.} In Table~\ref{tab:strategies}, GeOPD achieves the highest average score at both scales, surpassing SFT by 1.68 points at 4B and 0.98 at 9B, with wider margins over GRPO. At 4B, GeOPD consistently outperforms SFT across both the four multiple-choice benchmarks (+1.55 points on average) and open-ended HRVQA (+2.25 points). At 9B, where base multiple-choice scores saturate via option heuristics, GeOPD retains the leading multiple-choice average while surpassing SFT by 4.35 points on open-ended HRVQA. Free of exploitable option priors, this open-ended gain directly confirms genuine visual grounding over format memorization.

Against the distillation baselines, crop-based teachers such as Vision-OPD and VAD, as well as evidence-guided EAD and S-OPD, fall behind SFT at both scales. In wide aerial scenes, tight target crops discard the external spatial references needed to resolve ambiguity (Figure~\ref{fig:motivation}), while generic evidence alignment in EAD and two-sided contrast in S-OPD struggle without full-scene reference anchors and decoupled target masking. By marking target and reference regions directly within the full scene and isolating target-only interventions, GeOPD preserves global layout cues and outperforms EAD and S-OPD by 4.16 and 3.76 points at 4B, with 2.79 and 2.37 points at 9B, while surpassing multi-scale RS-OPSD by 1.59 points at 4B and RP-OPSD by 1.30 points at 9B.

\subsection{Ablation Study}
\label{sec:ablations}

\begin{table}[tbp]
\centering\footnotesize
\captionsetup{skip=4pt}
\caption{\textbf{Teacher evidence and answer privilege.} The last four rows share selection and contrast (\textit{w/o ref.} masks both targets and references). \textit{Original + answer} uses uniform JS.}
\label{tab:ablation}
\setlength{\tabcolsep}{2pt}
\begin{tabularx}{\columnwidth}{@{}>{\raggedright\arraybackslash}Xcccccc@{}}\toprule
Training configuration & CHOICE & GEO & Omni & HRVQA & LHRS & Avg.\\\midrule
Original + answer & 80.98 & 50.02 & 54.83 & 44.15 & 77.83 & 61.56\\
Marked + answer & 81.46 & 50.30 & 54.70 & 47.75 & 80.14 & 62.87\\
Original + facts & 80.67 & 51.04 & 53.84 & 47.50 & 79.71 & 62.55\\
Marked + facts (w/o ref.) & 80.64 & 54.47 & 53.61 & 49.05 & 79.57 & 63.47\\
\rowcolor{cyan!12}Marked + facts & 81.39 & 52.85 & 55.84 & 49.20 & 79.71 & \textbf{63.80}\\\bottomrule
\end{tabularx}
\end{table}

\noindent\textbf{Contextual evidence for the teacher.} 
Region marks anchor target and reference regions in the full scene, while visual facts state underlying observations. In Table~\ref{tab:ablation}, \textit{Original + answer} (61.56\%) serves as the answer-privileged baseline. Holding the selection and contrast objectives fixed across the remaining rows isolates each component. Even without textual facts, \textit{Marked + answer} (62.87\%) surpasses SFT and all prior OPD baselines. Replacing the answer with visual facts (\textit{Marked + facts}) improves the mean by 0.93 points, and adding region marks (\textit{Original + facts} to \textit{Marked + facts}) yields a 1.25-point gain. Conversely, occluding spatial references alongside targets (\textit{w/o ref.}) lowers the mean to 63.47\% and degrades counting and relational subtasks, while supervising SFT directly with visual facts also hurts accuracy (Appendix~\docref{app:baselines}{D}).

\begin{table}[tbp]
\centering\footnotesize
\captionsetup{skip=4pt}
\caption{\textbf{Effect of correction selection and visual contrast.} Sel. and Vis. indicate whether each component is enabled.}
\label{tab:mechanismplan}
\setlength{\tabcolsep}{2pt}
\begin{tabularx}{\columnwidth}{@{}>{\centering\arraybackslash}X>{\centering\arraybackslash}Xcccccc@{}}\toprule
Sel. & Vis. & CHOICE & GEO & Omni & HRVQA & LHRS & Avg.\\\midrule
$\times$ & $\times$ & 81.54 & 49.77 & 56.16 & 44.75 & 78.99 & 62.24\\
$\times$ & $\checkmark$ & 80.48 & 50.67 & 54.38 & 42.30 & 79.13 & 61.39\\
$\checkmark$ & $\times$ & 81.49 & 52.73 & 54.07 & 48.50 & 79.13 & 63.18\\
\rowcolor{cyan!12}$\checkmark$ & $\checkmark$ & 81.39 & 52.85 & 55.84 & 49.20 & 79.71 & \textbf{63.80}\\\bottomrule
\end{tabularx}
\end{table}

\noindent\textbf{Correction selection and visual contrast.} 
Table~\ref{tab:mechanismplan} decouples the training objectives under the identical \textit{Marked + facts} teacher input. Supplying this full privilege under uniform distillation (\text{Sel.} = $\times$, \text{Vis.} = $\times$) yields 62.24\%, matching RS-OPSD (62.21\%) and SFT (62.12\%) and confirming that stronger teacher privilege alone does not explain GeOPD's advantage. Applying visual contrast without directional filtering (\text{Sel.} = $\times$, \text{Vis.} = $\checkmark$) lowers the mean by 0.85 points, as the student imitates ungrounded tokens triggered by $e_{\text{fact}}$ and $\mathcal{L}_{\text{vis}}$ amplifies them. Enabling directional selection (\text{Sel.} = $\checkmark$, \text{Vis.} = $\times$) filters out non-visual corrections (+0.94 points), establishing grounded trajectories on which $\mathcal{L}_{\text{vis}}$ sharpens evidence sensitivity to reach 63.80\%.

\begin{table}[bp]
\centering\footnotesize
\captionsetup{skip=4pt}
\caption{\textbf{Correction weighting with a matched JS control.} Teacher inputs and student visual contrast are fixed.}
\label{tab:jsmatch}
\setlength{\tabcolsep}{2pt}
\begin{tabularx}{\columnwidth}{@{}>{\raggedright\arraybackslash}Xcccccc@{}}\toprule
Weighting rule & CHOICE & GEO & Omni & HRVQA & LHRS & Avg.\\\midrule
Uniform & 80.48 & 50.67 & 54.38 & 42.30 & 79.13 & 61.39\\
Uniform (matched JS) & 81.15 & 51.07 & 54.01 & 49.35 & 80.14 & 63.15\\
\rowcolor{cyan!12}Directional (GeOPD) & 81.39 & 52.85 & 55.84 & 49.20 & 79.71 & \textbf{63.80}\\\bottomrule
\end{tabularx}
\end{table}

\noindent\textbf{Correction allocation at matched strength.} 
Because directional weighting downweights unaligned tokens, Table~\ref{tab:jsmatch} tests whether its benefit comes simply from a smaller overall regularization strength. Scaling the batch JS loss in \textit{Uniform (matched JS)} to match the average magnitude of directional weighting recovers part of the gap by curbing over-regularization on ungrounded tokens. However, token-level allocation in \textit{Directional (GeOPD)} still outperforms this magnitude-matched control by 0.65 points on average and by nearly 2 points on complex multi-object scenes in GEOBench-VLM and OmniEarth, confirming the need to route supervision specifically to visually responsive tokens (see Appendix~\docref{app:baselines}{D} for sensitivity analysis on $\lambda$).

\begin{figure}[tbp]
\centering
\includegraphics[width=\columnwidth]{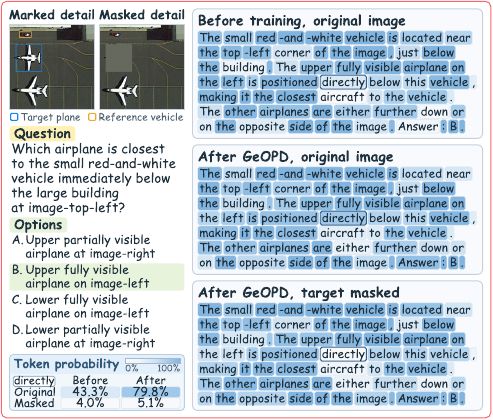}
\captionsetup{skip=4pt}
\caption{\textbf{Token probabilities under target masking.} Blue shading shows token probabilities from rescoring the same full response at fixed prefixes. The inset compares \emph{directly} across checkpoints and views. Square crops show the marked and masked views.}
\label{fig:token-response}
\end{figure}

\subsection{Task and Evidence Analysis}

\noindent\textbf{Visualizing Evidence Sensitivity.} The development question in Figure~\ref{fig:token-response} asks which airplane is closest to a reference vehicle. The airplane's position relative to that vehicle provides evidence for the answer. We compare the initial and trained 4B student by rescoring the trained student's response at fixed prefixes. With both objects visible, the probability of \emph{directly} in \emph{directly below} rises from 43.3\% to 79.8\% after training. With the airplane masked and the vehicle still visible, it changes from 4.0\% to 5.1\%. The larger shift on the original image shows that training sharpens the student's sensitivity to the airplane's visibility.

\begin{figure}[tbp]
\centering
\includegraphics[width=\columnwidth]{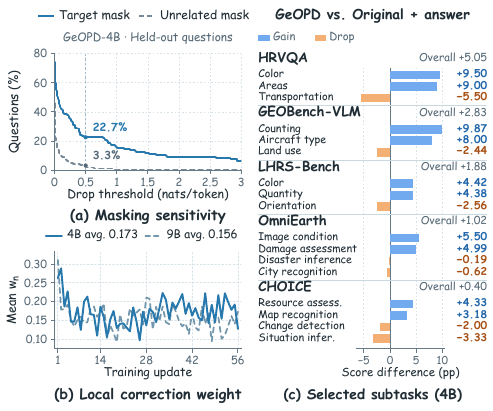}
\captionsetup{skip=4pt}
\caption{\textbf{Evidence sensitivity and subtask transfer.} (a) Share of held-out questions with at least the indicated drop in reference answer log probability per token from the original image. (b) Mean direction weight $w_n$ per training update. (c) Subtask score differences between GeOPD and the Original + answer baseline at 4B (in percentage points).}
\label{fig:training-transfer}
\end{figure}

\noindent\textbf{Evidence masking and correction weights.} 
On held-out questions, GeOPD-4B's reference answer scores drop substantially more under target masking than under matched unrelated masking across all thresholds while spatial references remain visible (Figure~\ref{fig:training-transfer}(a)), and occluding spatial references alone on relational items similarly causes a sharp drop, confirming reliance on both scene elements (Appendix~\docref{app:analysis}{E}). During training, GeOPD upweights teacher corrections aligned with the student's distributional shift under target restoration. Within a full response, only a fraction of tokens directly encodes target attributes or spatial relations, whereas remaining tokens form syntactic structure or refer to unoccluded landmarks. Consequently, the trajectory-averaged weights in Figure~\ref{fig:training-transfer}(b) remain stable at a selective level throughout training for both backbones, showing that directional selection consistently filters out unaligned or syntactic tokens while focusing supervision on visually decisive tokens such as \textit{directly} in Figure~\ref{fig:token-response}.

\noindent\textbf{Transfer across tasks.} 
Comparing GeOPD against the \textit{Original + answer} baseline in Figure~\ref{fig:training-transfer}(c) reveals how grounded visual supervision reshapes downstream capabilities compared to direct answer privilege. Supervision from marked scenes and visual facts substantially strengthens tasks that require inspecting fine-grained attributes within bounded regions, including object counting, color recognition across open and closed formats, and post-disaster damage assessment. Conversely, providing the raw reference answer retains a slight shortcut advantage on abstract situation inference or canonical orientation tasks that rely more on semantic priors than local visual contrast. Overall, grounding the teacher in scene evidence yields consistent net gains across all five benchmarks. Appendix~\docref{app:analysis}{E} details the full subtask breakdowns and masking controls.

\section{Conclusion}
In this work, we investigate how contextual evidence supports remote sensing scene understanding. To advance this task, we introduce GeoContext-5K, a training dataset that links questions and answers to spatial cues, visual facts, and supporting regions. Building on these annotations, we propose GeOPD to distill guidance from a frozen teacher that sees the marked full image and a visual fact. The method prioritizes corrections aligned with the student’s response to restored evidence and weights contrast between the student’s original and masked scores by the teacher’s sensitivity to masking. Across five remote sensing VQA benchmarks, GeOPD surpasses the evaluated open-weight MLLM baselines in mean score and leads the post-training comparison at each backbone size. At inference, the student answers from the original image and question alone. We hope this work encourages the use of contextual evidence as supervision across remote sensing tasks.

\section*{LLM Usage Statement}
\label{sec:llmusage}
During the construction of the dataset, we used the gpt-6.1-sol API to generate VQA pairs and label the supporting evidence, and employed the gemini-3.8-flash API to assist in the quality assessment and screening of the dataset by humans. OpenAI Codex helped with the code development and debugging, analysis of experimental results, writing and revision of the paper, as well as the creation of charts. The authors are responsible for the submitted content.

{\small\bibliographystyle{ieeenat_fullname}\bibliography{references}}

@inproceedings{wang2024earthvqa,
  title={{EarthVQA}: Towards queryable earth via relational reasoning-based remote sensing visual question answering},
  author={Wang, Junjue and Zheng, Zhuo and Chen, Zihang and Ma, Ailong and Zhong, Yanfei},
  booktitle={Proceedings of the AAAI conference on artificial intelligence},
  volume={38},
  number={6},
  pages={5481--5489},
  year={2024}
}

@inproceedings{hudson2019gqa,
  title={{GQA}: A new dataset for real-world visual reasoning and compositional question answering},
  author={Hudson, Drew A and Manning, Christopher D},
  booktitle={2019 IEEE/CVF Conference on Computer Vision and Pattern Recognition (CVPR)},
  pages={6693--6702},
  year={2019},
  organization={IEEE}
}

@inproceedings{wang2025xlrs,
  title={{XLRS-Bench}: Could your multimodal {LLMs} understand extremely large ultra-high-resolution remote sensing imagery?},
  author={Wang, Fengxiang and Wang, Hongzhen and Guo, Zonghao and Wang, Di and Wang, Yulin and Chen, Mingshuo and Ma, Qiang and Lan, Long and Yang, Wenjing and Zhang, Jing and others},
  booktitle={2025 IEEE/CVF Conference on Computer Vision and Pattern Recognition (CVPR)},
  pages={14325--14336},
  year={2025},
  organization={IEEE}
}

@InProceedings{pmlr-v267-shabbir25a,
  title = 	 {{G}eo{P}ixel: Pixel Grounding Large Multimodal Model in Remote Sensing},
  author =       {Shabbir, Akashah and Zumri, Mohammed and Bennamoun, Mohammed and Khan, Fahad Shahbaz and Khan, Salman},
  booktitle = 	 {Proceedings of the 42nd International Conference on Machine Learning},
  pages = 	 {54095--54111},
  year = 	 {2025},
  editor = 	 {Singh, Aarti and Fazel, Maryam and Hsu, Daniel and Lacoste-Julien, Simon and Berkenkamp, Felix and Maharaj, Tegan and Wagstaff, Kiri and Zhu, Jerry},
  volume = 	 {267},
  series = 	 {Proceedings of Machine Learning Research},
  month = 	 {13--19 Jul},
  publisher =    {PMLR},
  url = 	 {https://proceedings.mlr.press/v267/shabbir25a.html}
}

@inproceedings{xin2026segearth,
  title={{SegEarth-R2}: Towards comprehensive language-guided segmentation for remote sensing images},
  author={Xin, Zepeng and Li, Kaiyu and Chen, Luodi and Li, Wanchen and Yuchen, Xiao and Qiao, Hui and Zhang, Weizhan and Meng, Deyu and Cao, Xiangyong},
  booktitle={Proceedings of the IEEE/CVF Conference on Computer Vision and Pattern Recognition},
  pages={13199--13210},
  year={2026}
}

@inproceedings{zhang2026geovis,
  title={{GeoViS}: Geospatially rewarded visual search for remote sensing visual grounding},
  author={Zhang, Peirong and Zhang, Yidan and Xu, Luxiao and Lin, Jinliang and Guo, Zonghao and Wang, Fengxiang and Yang, Xue and Wei, Kaiwen and Wang, Lei},
  booktitle={Proceedings of the IEEE/CVF Conference on Computer Vision and Pattern Recognition},
  pages={14335--14345},
  year={2026}
}

@inproceedings{liu2026towards,
  title={Towards faithful reasoning in remote sensing: A perceptually-grounded geospatial chain-of-thought for vision-language models},
  author={Liu, Jiaqi and Sun, Lang and Fu, Ronghao and Yang, Bo},
  booktitle={International Conference on Learning Representations},
  volume={2026},
  pages={6535--6563},
  year={2026}
}

@inproceedings{gu2024minillm,
  title={{MiniLLM}: Knowledge distillation of large language models},
  author={Gu, Yuxian and Dong, Li and Wei, Furu and Huang, Minlie},
  booktitle={International Conference on Learning Representations},
  volume={2024},
  pages={32694--32717},
  year={2024}
}

@inproceedings{niu2021counterfactual,
  title={Counterfactual {VQA}: A cause-effect look at language bias},
  author={Niu, Yulei and Tang, Kaihua and Zhang, Hanwang and Lu, Zhiwu and Hua, Xian-Sheng and Wen, Ji-Rong},
  booktitle={2021 IEEE/CVF Conference on Computer Vision and Pattern Recognition (CVPR)},
  pages={12695--12705},
  year={2021},
  organization={IEEE}
}

@inproceedings{leng2024mitigating,
  title={Mitigating object hallucinations in large vision-language models through visual contrastive decoding},
  author={Leng, Sicong and Zhang, Hang and Chen, Guanzheng and Li, Xin and Lu, Shijian and Miao, Chunyan and Bing, Lidong},
  booktitle={2024 IEEE/CVF Conference on Computer Vision and Pattern Recognition (CVPR)},
  pages={13872--13882},
  year={2024},
  organization={IEEE}
}

@article{wang2026cv,
  title={{CV-OPD}: Contrastive Vision On-policy Distillation},
  author={Wang, Haowei and Lin, Hao and Zhu, Qi and Ji, Jiayi and Ji, Deyi and Cao, Liujuan and Ji, Rongrong},
  journal={arXiv preprint arXiv:2610.13264},
  year={2026}
}

@article{feng2026vc,
  title={{VC-OPD}: Visual Counterfactual On-Policy Distillation for Grounded Vision-Language Reasoning},
  author={Feng, Jiahe},
  journal={arXiv preprint arXiv:2610.14939},
  year={2026}
}

@article{xue2026distill,
  title={Distill What the Student Can See: Fisher-Projected On-Policy Distillation for Vision-Language Models},
  author={Xue, Leyan and Xiong, Feng and Ma, Mingjun and Zhang, Changqing},
  journal={arXiv preprint arXiv:2608.01263},
  year={2026}
}

@article{wei2026policy,
  title={On-Policy Visual Evidence Distillation},
  author={Wei, Shaohang and Song, Feifan and Peng, Guangyue and Yu, Wenhao and Li, Wei and Luo, Wen and Xu, Yang and Shen, Yufan and Mao, Luke and Du, Yang and others},
  journal={arXiv preprint arXiv:2609.36838},
  year={2026}
}

@article{qian2026med,
  title={Med-OPD: Improving Medical Vision-Language Models via Evidence-Aware On-Policy Distillation},
  author={Qian, Yunhang and Yu, Jiaquan and Liu, Jiawei and Wang, Meng and Li, Hongwei Bran and Hu, Xiaobin},
  journal={arXiv preprint arXiv:2607.16303},
  year={2026}
}

@article{tian2026vicur,
  title={Vicur: Visual cues as recoverable privilege for multimodal on-policy distillation},
  author={Tian, Kanghui and Liu, Siyuan and Yan, Ziang and Xia, Sheng and Dong, Shuai and Wang, Yi},
  journal={arXiv preprint arXiv:2606.05718},
  year={2026}
}

@article{sirko2026opd,
  title={Where-OPD: Spatially Guided On-Policy Self-Distillation of MLLMs with Synthetic Scenes},
  author={Sirko-Galouchenko, Sophia and Wysoczanska, Monika and Bursuc, Andrei and Thome, Nicolas and Gidaris, Spyros},
  journal={arXiv preprint arXiv:2610.02117},
  year={2026}
}

@article{jiang2026rs,
  title={RS-OPSD: Reliable Privileged On-Policy-Self-Distillation for Ultra-High-Resolution Remote Sensing VQA},
  author={Jiang, Chengjie and Zhou, Yunqi and Yan, Jiafeng and Zhao, Sihang and Yuan, Chun and Li, Jing},
  journal={arXiv preprint arXiv:2609.38072},
  year={2026}
}

@article{yuan2026vision,
  title={Vision-opd: Learning to see fine details for multimodal llms via on-policy self-distillation},
  author={Yuan, Qianhao and Lou, Jie and Yu, Xing and Lin, Hongyu and Sun, Le and Han, Xianpei and Lu, Yaojie},
  journal={arXiv preprint arXiv:2605.18740},
  year={2026}
}

@article{hu2026evidence,
  title={Evidence-Aligned Multimodal On-Policy Self-Distillation for Fine-Grained Visual Understanding},
  author={Hu, Nanxing and Yan, Qiwei and Zhang, Jinchao and Kang, Guoliang},
  journal={arXiv preprint arXiv:2609.34672},
  year={2026}
}

@article{liu2026visual,
  title={Visual-advantage on-policy distillation for vision-language models},
  author={Liu, Ruiqi and Lv, Xiaolei and Li, Gengsheng and Zhu, Ximo and Wang, Zhiheng and Zhang, Zhengbo and Chen, Junkai and Li, Zhiheng and Li, Bo and Gao, Jun and others},
  journal={arXiv preprint arXiv:2605.21924},
  year={2026}
}

@article{zhang2026vad,
  title={Vad: Attributing visual evidence for target reconstruction in multimodal on-policy distillation},
  author={Zhang, Kangning and Li, Yixing and Shao, Shuai and Li, Qingyao and Lu, Zhengxi and Yao, Zhiyuan and Lin, Jianghao and Jiao, Wenxiang and Lu, Yuan and Liu, Weiwen and others},
  journal={arXiv preprint arXiv:2607.28590},
  year={2026}
}

@article{li2024vrsbench,
  title={{VRSBench}: A versatile vision-language benchmark dataset for remote sensing image understanding},
  author={Li, Xiang and Ding, Jian and Elhoseiny, Mohamed},
  journal={Advances in Neural Information Processing Systems},
  volume={37},
  pages={3229--3242},
  year={2024}
}

@article{zhang2024rs5m,
  title={{RS5M} and {GeoRSCLIP}: A large-scale vision-language dataset and a large vision-language model for remote sensing},
  author={Zhang, Zilun and Zhao, Tiancheng and Guo, Yulong and Yin, Jianwei},
  journal={IEEE Transactions on Geoscience and Remote Sensing},
  volume={62},
  pages={1--23},
  year={2024},
  publisher={IEEE}
}

@inproceedings{liu2026zoomearth,
  title={{ZoomEarth}: Active perception for ultra-high-resolution geospatial vision-language tasks},
  author={Liu, Ruixun and Fu, Bowen and Song, Jiayi and Li, Kaiyu and Li, Wanchen and Xue, Lanxuan and Qiao, Hui and Zhang, Weizhan and Meng, Deyu and Cao, Xiangyong},
  booktitle={Proceedings of the IEEE/CVF Conference on Computer Vision and Pattern Recognition},
  pages={34877--34888},
  year={2026}
}

@article{an2026choice,
  title={{CHOICE}: Benchmarking the Remote Sensing Capabilities of Large Vision-Language Models},
  author={An, Xiao and Sun, Jiaxing and Gui, Zihan and He, Wei},
  journal={Advances in Neural Information Processing Systems},
  volume={38},
  year={2025}
}

@inproceedings{danish2025geobench,
  title={{GEOBench-VLM}: Benchmarking vision-language models for geospatial tasks},
  author={Danish, Muhammad and Munir, Muhammad Akhtar and Shah, Syed Roshaan Ali and Kuckreja, Kartik and Khan, Fahad Shahbaz and Fraccaro, Paolo and Lacoste, Alexandre and Khan, Salman},
  booktitle={Proceedings of the IEEE/CVF International Conference on Computer Vision},
  pages={7132--7142},
  year={2025}
}

@article{fu2026omniearth,
  title={{OmniEarth}: A Benchmark for Evaluating Vision-Language Models in Geospatial Tasks},
  author={Fu, Ronghao and Liu, Haoran and Zhang, Weijie and Lin, Zhiwen and Yang, Xiao and Zhang, Peng and Yang, Bo},
  journal={arXiv preprint arXiv:2603.09471},
  year={2026}
}

@inproceedings{agarwal2024policy,
  title={On-policy distillation of language models: Learning from self-generated mistakes},
  author={Agarwal, Rishabh and Vieillard, Nino and Zhou, Yongchao and Stanczyk, Piotr and Ramos Garea, Sabela and Geist, Matthieu and Bachem, Olivier},
  booktitle={International Conference on Learning Representations},
  volume={2024},
  pages={21246--21263},
  year={2024}
}

@article{lobry2020rsvqa,
  title={RSVQA: Visual question answering for remote sensing data},
  author={Lobry, Sylvain and Marcos, Diego and Murray, Jesse and Tuia, Devis},
  journal={IEEE Transactions on Geoscience and Remote Sensing},
  volume={58},
  number={12},
  pages={8555--8566},
  year={2020},
  publisher={IEEE}
}

@article{chen2024expanding,
  title={Expanding performance boundaries of open-source multimodal models with model, data, and test-time scaling},
  author={Chen, Zhe and Wang, Weiyun and Cao, Yue and Liu, Yangzhou and Gao, Zhangwei and Cui, Erfei and Zhu, Jinguo and Ye, Shenglong and Tian, Hao and Liu, Zhaoyang and others},
  journal={arXiv preprint arXiv:2412.05271},
  year={2024}
}

@article{wang2025internvl3,
  title={Internvl3.5: Advancing open-source multimodal models in versatility, reasoning, and efficiency},
  author={Wang, Weiyun and Gao, Zhangwei and Gu, Lixin and Pu, Hengjun and Cui, Long and Wei, Xingguang and Liu, Zhaoyang and Jing, Linglin and Ye, Shenglong and Shao, Jie and others},
  journal={arXiv preprint arXiv:2508.18265},
  year={2025}
}

@misc{bai2025qwen25vltechnicalreport,
      title={Qwen2.5-VL Technical Report}, 
      author={Shuai Bai and Keqin Chen and Xuejing Liu and Jialin Wang and Wenbin Ge and Sibo Song and Kai Dang and Peng Wang and Shijie Wang and Jun Tang and Humen Zhong and Yuanzhi Zhu and Mingkun Yang and Zhaohai Li and Jianqiang Wan and Pengfei Wang and Wei Ding and Zheren Fu and Yiheng Xu and Jiabo Ye and Xi Zhang and Tianbao Xie and Zesen Cheng and Hang Zhang and Zhibo Yang and Haiyang Xu and Junyang Lin},
      year={2025},
      eprint={2502.13923},
      archivePrefix={arXiv},
      primaryClass={cs.CV},
      url={https://arxiv.org/abs/2502.13923}, 
}

@InProceedings{Yu_2026_CVPR,
    author    = {Yu, Tianyu and Wang, Zefan and Wang, Chongyi and Huang, Fuwei and Ma, Wenshuo and He, Zhihui and Cai, Tianchi and Chen, Weize and Huang, Yuxiang and Zhao, Ranchi and Xu, Bokai and Cui, Junbo and Xu, Yingjing and Ruan, Liqing and Zhang, Luoyuan and Liu, Hanyu and Tang, Jingkun and Liu, Hongyuan and Guo, Qining and Hu, Wenhao and He, Bingxiang and Zhou, Jie and Cai, Jie and Qi, Ji and Guo, Zonghao and Chen, Chi and Zeng, Guoyang and Li, Yuxuan and Cui, Ganqu and Ding, Ning and Han, Xu and Yao, Yuan and Liu, Zhiyuan and Sun, Maosong},
    title     = {MiniCPM-V 4.5: Cooking Efficient MLLMs via Architecture, Data, and Training Recipe},
    booktitle = {Proceedings of the IEEE/CVF Conference on Computer Vision and Pattern Recognition (CVPR)},
    month     = {June},
    year      = {2026},
    pages     = {11704-11715}
}

@article{yue2025mimo,
  title={Mimo-vl technical report},
  author={Yue, Zihao and Lin, Zhenru and Song, Yifan and Wang, Weikun and Ren, Shuhuai and Gu, Shuhao and Li, Shicheng and Li, Peidian and Zhao, Liang and Li, Lei and others},
  journal={arXiv preprint arXiv:2506.03569},
  year={2025}
}

@article{hong2025glm,
  title={Glm-4.5 v and glm-4.1 v-thinking: Towards versatile multimodal reasoning with scalable reinforcement learning},
  author={Hong, Wenyi and Yu, Wenmeng and Gu, Xiaotao and Wang, Guo and Gan, Guobing and Tang, Haomiao and Cheng, Jiale and Qi, Ji and Ji, Junhui and Pan, Lihang and others},
  journal={arXiv preprint arXiv:2507.01006},
  year={2025}
}

@InProceedings{pmlr-v306-wei26j,
  title = 	 {Zooming without Zooming: Region-to-Image Distillation for Fine-Grained Multimodal Perception},
  author =       {Wei, Lai and He, Liangbo and Lan, Jun and Dong, Lingzhong and Cai, Yutong and Li, Siyuan and Zhu, Huijia and Wang, Weiqiang and Kong, Linghe and Wang, Yue and Zhang, Zhuosheng and Huang, Weiran},
  booktitle = 	 {Proceedings of the 43rd International Conference on Machine Learning},
  pages = 	 {133660--133688},
  year = 	 {2026},
  editor = 	 {Zhang, Tong and Dudik, Miroslav and Jaggi, Martin and Agarwal, Alekh and Li, Sharon and Schuurmans, Dale and Zhu, Jerry and Berkenkamp, Felix and Dong, Hanze and Bietti, Alberto},
  volume = 	 {306},
  series = 	 {Proceedings of Machine Learning Research},
  month = 	 {06--11 Jul},
  publisher =    {PMLR},
  url = 	 {https://proceedings.mlr.press/v306/wei26j.html}
}

@article{bai2025qwen3,
  title={Qwen3-vl technical report},
  author={Bai, Shuai and Cai, Yuxuan and Chen, Ruizhe and Chen, Keqin and Chen, Xionghui and Cheng, Zesen and Deng, Lianghao and Ding, Wei and Gao, Chang and Ge, Chunjiang and others},
  journal={arXiv preprint arXiv:2511.21631},
  year={2025}
}

@misc{qwen3.5,
    title  = {{Qwen3.5}: Towards Native Multimodal Agents},
    author = {{Qwen Team}},
    month  = {February},
    year   = {2026},
    url    = {https://qwen.ai/blog?id=qwen3.5}
}

@misc{openai2026gpt61sol,
    title        = {{GPT-6.1 Sol} Model},
    author       = {{OpenAI}},
    howpublished = {\href{https://developers.openai.com/api/docs/models/gpt-6.1-sol}{OpenAI API documentation}},
    year         = {2026},
    url          = {https://developers.openai.com/api/docs/models/gpt-6.1-sol},
    note         = {Accessed November 10, 2026}
}

@inproceedings{pang2025vhm,
  title={Vhm: Versatile and honest vision language model for remote sensing image analysis},
  author={Pang, Chao and Weng, Xingxing and Wu, Jiang and Li, Jiayu and Liu, Yi and Sun, Jiaxing and Li, Weijia and Wang, Shuai and Feng, Litong and Xia, Gui-Song and others},
  booktitle={Proceedings of the AAAI Conference on Artificial Intelligence},
  volume={39},
  number={6},
  pages={6381--6388},
  year={2025}
}

@inproceedings{yao2026remotereasoner,
  title={Remotereasoner: Towards unifying geospatial reasoning workflow},
  author={Yao, Liang and Liu, Fan and Lu, Hongbo and Zhang, Chuanyi and Min, Rui and Xu, Shengxiang and Di, Shimin and Peng, Pai},
  booktitle={Proceedings of the AAAI Conference on Artificial Intelligence},
  volume={40},
  number={14},
  pages={11883--11891},
  year={2026}
}

@inproceedings{kuckreja2024geochat,
  title={Geochat: Grounded large vision-language model for remote sensing},
  author={Kuckreja, Kartik and Danish, Muhammad Sohail and Naseer, Muzammal and Das, Abhijit and Khan, Salman and Khan, Fahad Shahbaz},
  booktitle={2024 IEEE/CVF Conference on Computer Vision and Pattern Recognition (CVPR)},
  pages={27831--27840},
  year={2024},
  organization={IEEE}
}

@InProceedings{Shu_2026_CVPR,
    author    = {Shu, Yan and Ren, Bin and Xiong, Zhitong and Zhu, Xiao Xiang and Demir, Beg\"um and Sebe, Nicu and Rota, Paolo},
    title     = {TerraScope: Pixel-Grounded Visual Reasoning for Earth Observation},
    booktitle = {Proceedings of the IEEE/CVF Conference on Computer Vision and Pattern Recognition (CVPR)},
    month     = {June},
    year      = {2026},
    pages     = {16712-16722}
}

@article{zhu2026rp,
  title={Rp-opsd: Resolution-privileged on-policy self-distillation for multimodal large language models},
  author={Zhu, Qihui and Wang, Yuchen and Wen, Zijian and Zhang, Tao and Zhang, Mengjie and Liu, Yang and Chen, Shuangwu and Wu, Siying and Yang, Jian and Jiang, Xiaofeng},
  journal={arXiv preprint arXiv:2607.24447},
  year={2026}
}

@article{liu2026better,
  title={Is Better Teacher Supervision Enough? Unlocking Student-side Learning in Multimodal On-Policy Distillation},
  author={Liu, Siyuan and Tian, Kanghui and Duan, Yue and He, Yutao and Yang, Shangdong and Zhang, Jian and Shi, Yinghuan},
  journal={arXiv preprint arXiv:2609.39120},
  year={2026}
}

@article{luo2024skysensegpt,
  title={Skysensegpt: A fine-grained instruction tuning dataset and model for remote sensing vision-language understanding},
  author={Luo, Junwei and Pang, Zhen and Zhang, Yongjun and Wang, Tingzhu and Wang, Linlin and Dang, Bo and Lao, Jiangwei and Wang, Jian and Chen, Jingdong and Tan, Yihua and others},
  journal={arXiv preprint arXiv:2406.10100},
  year={2024}
}

@article{shu2025earthmind,
  title={Earthmind: Leveraging cross-sensor data for advanced earth observation interpretation with a unified multimodal llm},
  author={Shu, Yan and Ren, Bin and Xiong, Zhitong and Paudel, Danda Pani and Van Gool, Luc and Demir, Beg{\"u}m and Sebe, Nicu and Rota, Paolo},
  journal={arXiv preprint arXiv:2506.01667},
  year={2025}
}

@inproceedings{soni2025earthdial,
  title={Earthdial: Turning multi-sensory earth observations to interactive dialogues},
  author={Soni, Sagar and Dudhane, Akshay and Debary, Hiyam and Fiaz, Mustansar and Munir, Muhammad Akhtar and Danish, Muhammad Sohail and Fraccaro, Paolo and Watson, Campbell D and Klein, Levente J and Khan, Fahad Shahbaz and others},
  booktitle={2025 IEEE/CVF Conference on Computer Vision and Pattern Recognition (CVPR)},
  pages={14303--14313},
  year={2025},
  organization={IEEE}
}

@article{li2024hrvqa,
  title={{HRVQA}: A visual question answering benchmark for high-resolution aerial images},
  author={Li, Kun and Vosselman, George and Yang, Michael Ying},
  journal={ISPRS Journal of Photogrammetry and Remote Sensing},
  volume={214},
  pages={65--81},
  year={2024},
  publisher={Elsevier}
}

@inproceedings{muhtar2024lhrs,
  title={{LHRS-Bot}: Empowering remote sensing with {VGI}-enhanced large multimodal language model},
  author={Muhtar, Dilxat and Li, Zhenshi and Gu, Feng and Zhang, Xueliang and Xiao, Pengfeng},
  booktitle={European Conference on Computer Vision},
  pages={440--457},
  year={2024},
  organization={Springer}
}

@inproceedings{liu2023visual,
  title={Visual Instruction Tuning},
  author={Liu, Haotian and Li, Chunyuan and Wu, Qingyang and Lee, Yong Jae},
  booktitle={Advances in Neural Information Processing Systems},
  volume={36},
  pages={34892--34916},
  year={2023},
  doi={10.52202/075280-1516}
}

@article{shao2024deepseekmath,
  title={Deepseekmath: Pushing the limits of mathematical reasoning in open language models},
  author={Shao, Zhihong and Wang, Peiyi and Zhu, Qihao and Xu, Runxin and Song, Junxiao and Bi, Xiao and Zhang, Haowei and Zhang, Mingchuan and Li, YK and Wu, Yang and others},
  journal={arXiv preprint arXiv:2402.03300},
  year={2024}
}

@inproceedings{xia2023openearthmap,
  title={{OpenEarthMap}: A benchmark dataset for global high-resolution land cover mapping},
  author={Xia, Junshi and Yokoya, Naoto and Adriano, Bruno and Broni-Bediako, Clifford},
  booktitle={2023 IEEE/CVF Winter Conference on Applications of Computer Vision (WACV)},
  pages={6254--6264},
  year={2023},
  organization={IEEE}
}

@article{yuan2024rrsis,
  title={{RRSIS}: Referring remote sensing image segmentation},
  author={Yuan, Zhenghang and Mou, Lichao and Hua, Yuansheng and Zhu, Xiao Xiang},
  journal={IEEE Transactions on Geoscience and Remote Sensing},
  volume={62},
  pages={1--12},
  year={2024},
  publisher={IEEE}
}

\clearpage
\setcounter{figure}{0}
\setcounter{table}{0}
\setcounter{equation}{0}
\renewcommand{\theHfigure}{supp.\arabic{figure}}
\renewcommand{\theHtable}{supp.\arabic{table}}
\renewcommand{\theHequation}{supp.\arabic{equation}}
\maketitlesupplementary
\appendix
\raggedbottom
\renewcommand{\thefigure}{S\arabic{figure}}
\renewcommand{\thetable}{S\arabic{table}}
\renewcommand{\theequation}{S\arabic{equation}}

\section{GeoContext-5K Construction}
\label{app:data}

\subsection{Sources, Annotation, and Quality Checks}
GeoContext-5K contains 5,394 reviewed training QA pairs from 5,340 recorded source image groups. Figure~\docref{fig:data}{2} outlines the construction process.

\noindent\textbf{Source and task selection.}
We select tasks based on scene content, available spatial references, and the region needed to answer, using imagery from VRSBench~\cite{li2024vrsbench}, LRS-GRO~\cite{liu2026zoomearth}, and RS5M~\cite{zhang2024rs5m}. The training set includes 2,658 multiple-choice and 2,736 short-answer questions across six task families.

\noindent\textbf{Joint QA and evidence construction.}
We use gpt-6.1-sol~\cite{openai2026gpt61sol} to draft the question, answer, spatial cue, visual fact, and supporting regions together. The cue identifies where to look, and the fact records a visible observation supporting the answer. Boxes distinguish target evidence from spatial references or search boundaries, using full-scene coordinates.

\noindent\textbf{Independent review.}
Using gemini-3.8-flash, the review pipeline first records an independent answer from the unmarked image and the question, with options when applicable. A subsequent call checks answer equivalence, referent clarity, visibility at the student input scale, option consistency, and the correspondence between stated facts and marked regions. Human spot-checks supplement model review. Ambiguous questions are revised, and records without verifiable answers or evidence are excluded.

Table~\ref{tab:dataset-comparison} compares evidence inputs in documented privileged OPD protocols.

\begin{table*}[t]
\centering\small
\captionsetup{skip=4pt}
\caption{\textbf{Evidence used in privileged OPD training.}
Evidence boxes can be used by either model. Visual facts are recorded per question. Checkmarks indicate documented use, and dashes indicate use was not reported. Target masking applies to eligible questions.}
\label{tab:dataset-comparison}
\setlength{\tabcolsep}{5pt}
\renewcommand{\arraystretch}{1.14}
\begin{tabularx}{\textwidth}{@{}l r Y Y Y Y@{}}
\toprule
Method / training data & QA pairs & Global teacher view & Evidence boxes & Visual facts & Target masking\\
\midrule
Vision-OPD~\cite{yuan2026vision} & 6,241 & -- & $\checkmark$ & -- & --\\
RP-OPSD~\cite{zhu2026rp} & 5,295 & $\checkmark$ & -- & -- & --\\
RS-OPSD / GeoEvidence-6K~\cite{jiang2026rs} & 6,750 & $\checkmark$ & $\checkmark$ & -- & --\\
\textbf{GeOPD / GeoContext-5K} & \textbf{5,394} & \evidencecheck & \evidencecheck & \evidencecheck & \evidencecheck\\
\bottomrule
\end{tabularx}
\end{table*}

\subsection{Authoring and Review Prompts}
Figure~\ref{fig:annotation-prompts} shows the API instructions for annotation and review.

\begin{figure}[!t]
\begin{tcolorbox}[title={(a) Joint QA and evidence construction},
  colback=promptblue,colframe=cvprblue,fonttitle=\bfseries\small,
  boxrule=0.55pt,arc=1.4mm,left=4pt,right=4pt,top=3pt,bottom=3pt]
\small
Inspect the unmarked image at the student input scale. Write a visually answerable question with a unique referent and an explicit scope. For multiple choice, write four distinct options and one correct option. Use an empty options object for short answers. Record where to look in the spatial cue and the visible observations in the visual fact. Mark targets and references separately in normalized \texttt{xyxy} coordinates. Return JSON.
\begin{lstlisting}[basicstyle=\ttfamily\footnotesize,columns=fullflexible,keepspaces=true]
{
  "question": "",
  "options": {},
  "reference_answer": "",
  "spatial_cue": "",
  "visual_fact": "",
  "target_bboxes": [],
  "reference_bboxes": []
}
\end{lstlisting}
\end{tcolorbox}
\begin{tcolorbox}[title={(b) Blind answering and evidence review},
  colback=promptgreen,colframe=green!40!black,fonttitle=\bfseries\small,
  boxrule=0.55pt,arc=1.4mm,left=4pt,right=4pt,top=3pt,bottom=3pt]
\small
\textbf{Blind call.} Use the unmarked image, question, and any options to produce an independent answer. Save the response before seeing the annotation.
\par\smallskip
\textbf{Verification call.} Compare the saved answer with the proposed reference answer. Inspect whether the question has a unique referent, the fact is visible, and the target and reference boxes cover the described regions. Check option consistency and whether masking can preserve the spatial reference. Return a verdict of \texttt{pass}, \texttt{revise}, \texttt{uncertain}, or \texttt{fail}, with reasons.
\end{tcolorbox}
\captionsetup{skip=4pt}
\caption{\textbf{Prompt templates for GeoContext-5K.} Panel (a) shows the annotation prompt and output schema. Panel (b) shows the independent answer and evidence checks.}
\label{fig:annotation-prompts}
\end{figure}

\subsection{Annotated Examples}
\begin{figure*}[tp]
  \centering
  \includegraphics[width=\textwidth,height=0.93\textheight,keepaspectratio]{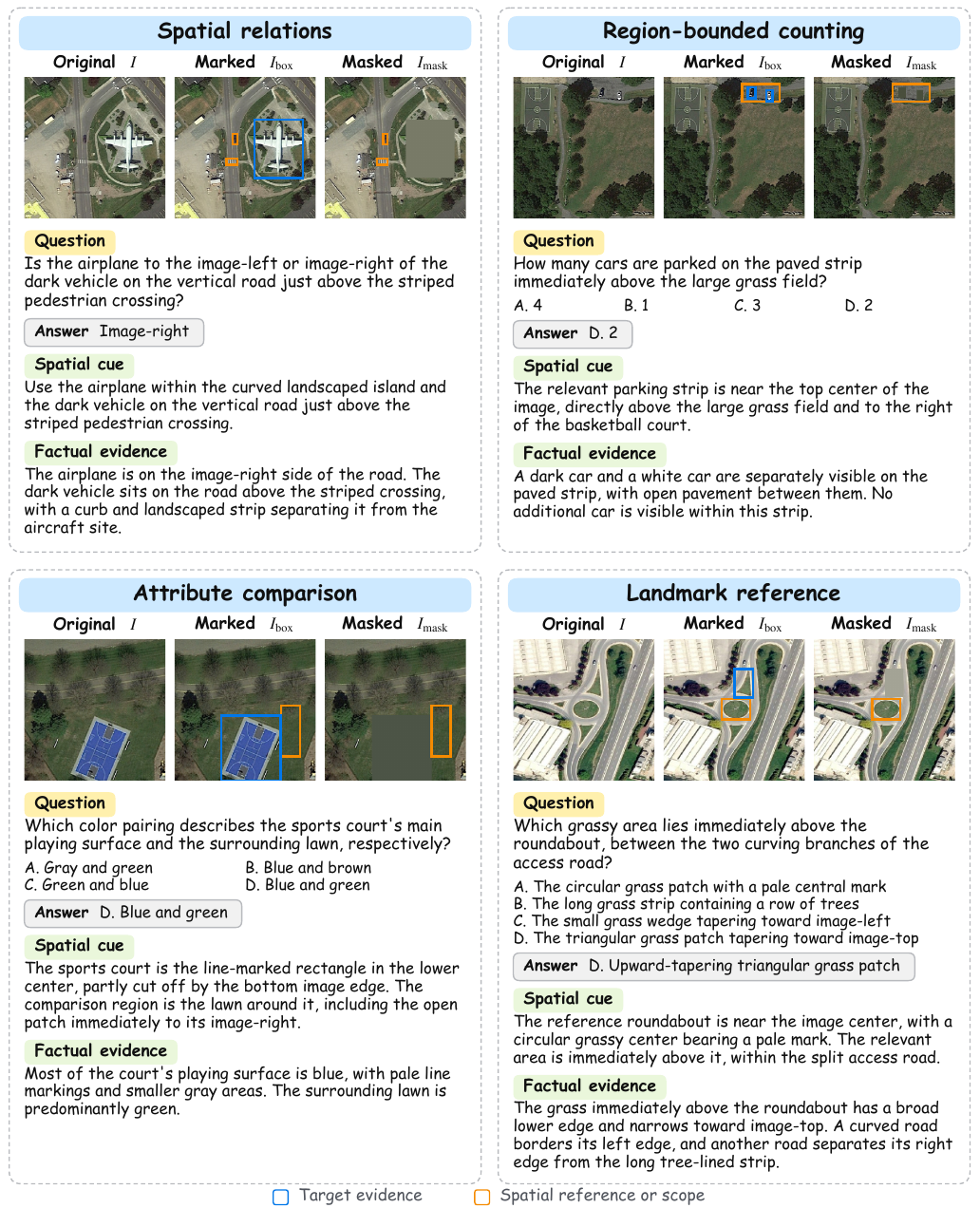}
  \captionsetup{skip=4pt}
  \caption{\textbf{GeoContext-5K annotations across four tasks.}
  Each triplet uses the same full image and coordinates. The marked view identifies the target in blue and its spatial reference or scope in orange. The masked view covers the target pixels and displays the reference outline for readability.}
  \label{fig:dataset-samples}
\end{figure*}

\begin{figure*}[tp]
  \centering
  \includegraphics[page=1,width=\textwidth,height=0.93\textheight,keepaspectratio]{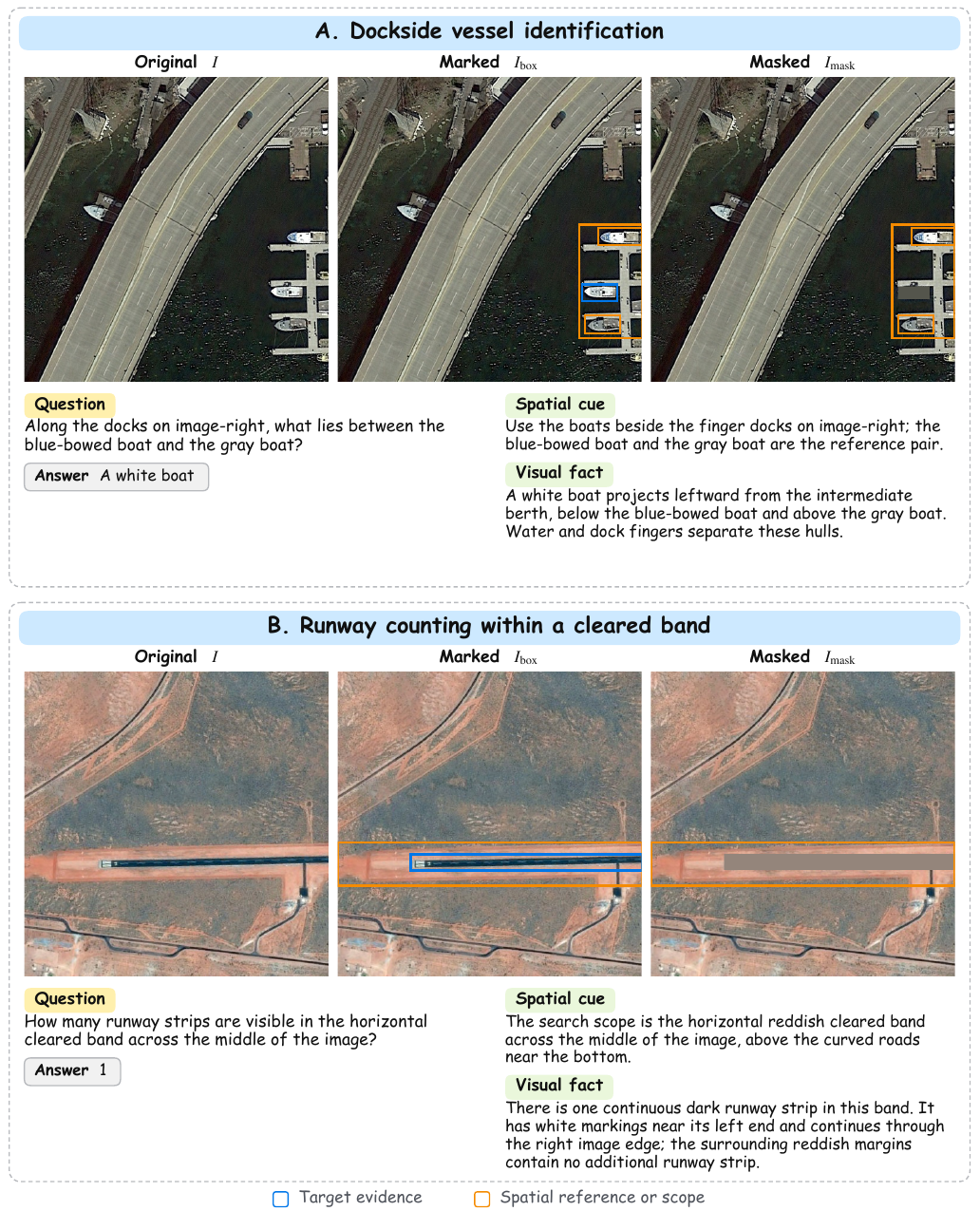}
  \captionsetup{skip=4pt}
  \caption{\textbf{Dockside identification and scoped counting.}
  The boats in adjacent berths define the search around the target hull in panel A. In panel B, the cleared band defines the scope for counting visible runway strips.}
  \label{fig:dataset-samples-scope}
\end{figure*}

\begin{figure*}[tp]
  \centering
  \includegraphics[page=2,width=\textwidth,height=0.93\textheight,keepaspectratio]{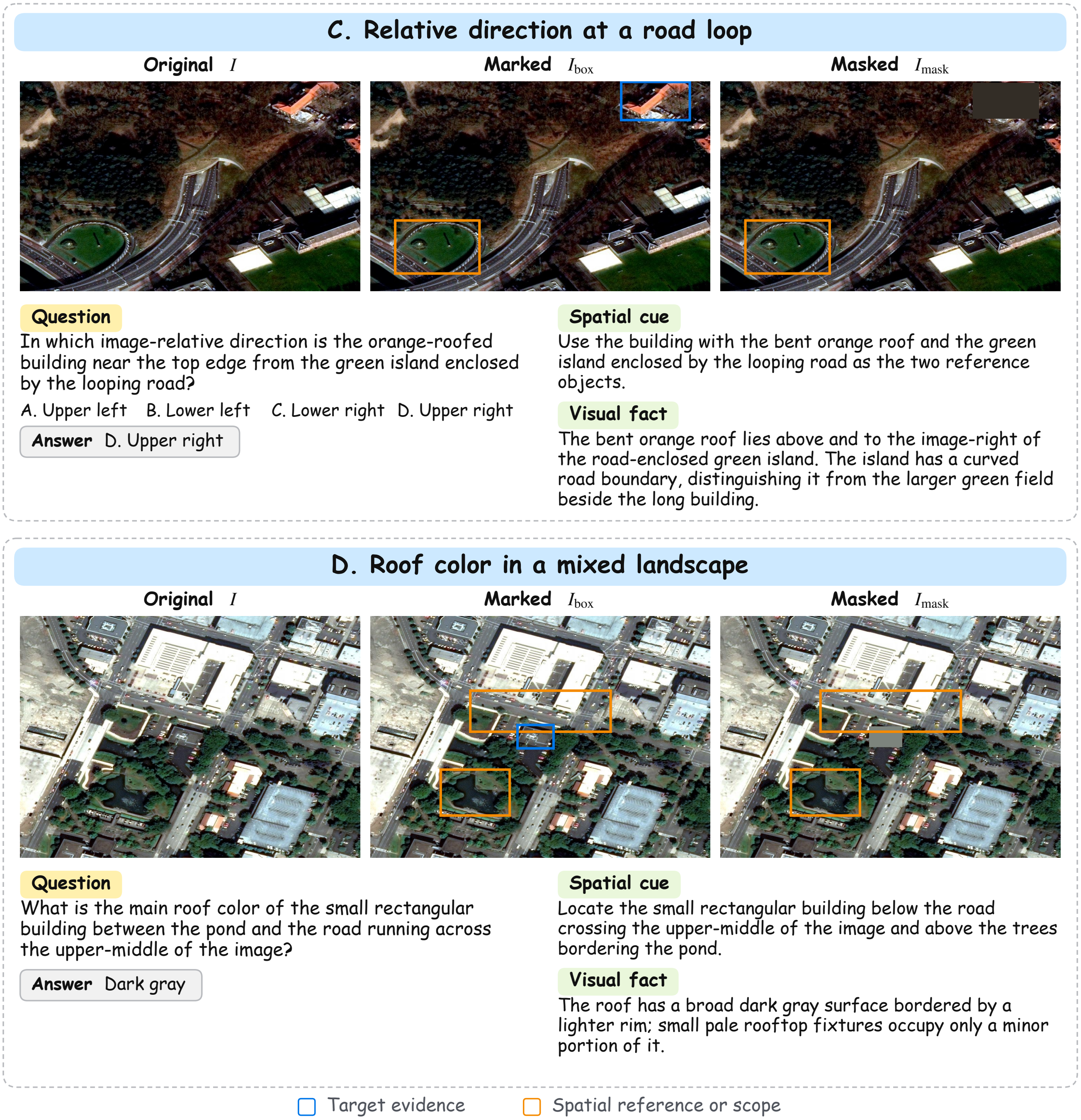}
  \captionsetup{skip=4pt}
  \caption{\textbf{Relative direction and contextual attributes.}
  In panel C, the island enclosed by the looping road provides the reference for the position of the orange-roofed building. In panel D, the pond and road distinguish the roof whose color is queried from nearby buildings.}
  \label{fig:dataset-samples-spatial}
\end{figure*}

Figures~\ref{fig:dataset-samples}--\ref{fig:dataset-samples-spatial} show eight training records. The masked view fills the target region with the original image's mean RGB color. Reference outlines in the masked views are for illustration only.

\section{Training and Computational Cost}
\label{app:implementation}

\subsection{Configuration and Inference}
\label{app:training-config}
\noindent\textbf{Training.}
We initialize the student and frozen teacher from the same Qwen3.5-4B or Qwen3.5-9B checkpoint~\cite{qwen3.5}. Both GeOPD runs use a batch of 96 questions, four sampled responses per question, and 56 updates over one epoch at a learning rate of $2\times10^{-6}$. They use four NVIDIA RTX PRO 6000 Blackwell GPUs, FSDP, BF16 computation with FP32 actor parameters, gradient clipping at 1.0, and weight decay of 0.01. Rollout tensor parallelism is one at 4B and two at 9B. The 4B actor uses a microbatch of one per GPU and no learning rate warmup. 

\noindent\textbf{Objective implementation.}
Directional distillation computes JS with a mixture coefficient of 0.5 on the 100 student tokens with the largest original-image probability and a residual bucket. The inverse-probability floor is $10^{-8}$, and direction comparisons with a norm product below $10^{-8}$ receive zero weight. Rollout importance correction uses a weight cap of 2. The original and masked views share the question, image dimensions, and multimodal positional layout. The masked score and the teacher passes are detached.

\noindent\textbf{Inference.}
Qwen checkpoints use the same prompts, RGB images with a long edge of at most 1,024 pixels, and greedy decoding with a 128-token response budget and an 8,192-token context limit. The processor allows 65,536 to 1,048,576 pixels. Other architectures use their supported processors. At inference, the GeOPD student answers from the original image and question.

\subsection{Logged Training Cost}
Table~\ref{tab:training-cost} reports the 4B runs. GeOPD reuses its sampled responses for teacher and masked-view scoring with gradients disabled on those auxiliary passes.

\begin{table*}[t]
\centering\small
\captionsetup{skip=4pt}
\caption{\textbf{Online 4B post-training cost on $4\times$ NVIDIA RTX PRO 6000 Blackwell GPUs.} Scoring forwards count logical model and view evaluations per fixed response, including the trainable student pass. The RS-OPSD probe is a separate teacher evaluation. Step time is averaged over 56 updates, GPU hours equal elapsed training hours times four, and peak memory is the maximum logged allocation.}
\label{tab:training-cost}
\setlength{\tabcolsep}{9pt}
\begin{tabular}{@{}lccrrr@{}}
\toprule
Method & Rollouts / question & Scoring forwards & Time (s/step) & Peak (GiB) & GPU hours\\
\midrule
GRPO~\cite{shao2024deepseekmath} & 4 & 2 & 430.07 & 43.85 & 26.76\\
OPSD~\cite{agarwal2024policy} & 4 & 2 & 610.09 & 50.36 & 37.96\\
RS-OPSD~\cite{jiang2026rs} & 4 & 3 $+$ probe & 675.65 & 57.33 & 42.04\\
GeOPD & 4 & 5 with contrast & 579.53 & 49.88 & 36.06\\
\bottomrule
\end{tabular}
\end{table*}

With both objectives active, GeOPD scores the student on both views, the privileged teacher, and the frozen teacher on the original and masked images. Measured runtime also depends on microbatching, rollout recomputation, and the cost of processing different image inputs.

\section{Evaluation Protocol}
\label{app:evaluation}

\subsection{Benchmarks and Scoring}
Table~\ref{tab:evalinventory} lists the evaluated task groups. The main-paper average weights the five benchmarks equally, and each benchmark score aggregates its evaluation records. Training efficiency and computational cost are detailed in Section~\ref{app:implementation}.

\noindent\textbf{Standardized evaluation protocol.}
We use one greedy decode per question, fixed option prompts, and the same answer extraction rules within each benchmark.

\begin{table*}[t]
\centering\footnotesize
\captionsetup{skip=4pt}
\caption{\textbf{Evaluated subtask inventory.} The five evaluation sets contain 20,964 records. Italics indicate the answer format.}
\label{tab:evalinventory}
\setlength{\tabcolsep}{3pt}
\renewcommand{\arraystretch}{1.18}
\begin{tabular*}{\textwidth}{@{\extracolsep{\fill}}>{\raggedright\arraybackslash}p{2.2cm}cc>{\raggedright\arraybackslash}p{11.9cm}@{}}\toprule
Benchmark & Groups/tags & Records & Included subtask types\\\midrule
CHOICE & 21 & 9,407 & \textit{MCQ.} Attribute comparison; Attribute recognition; Change detection; Disaster discrimination; Environmental assessment; Geospatial determination; Hallucination detection; Image caption; Image modality; Image quality; Landmark recognition; Map recognition; Object counting; Object localization; Object presence; Physical property; Resource assessment; Scene classification; Situation inference; Spatial relationship; Time property.\\
\addlinespace
GEOBench-VLM & 17 & 3,211 & \textit{MCQ.} Aircraft Type Classification; Building Counting; Crop Type Classification; Disaster Type Classification; Fire Risk Assessment; General Aircraft Counting; General Vehicle Counting; Land Use Classification; Marine debris counting; Scene Classification; Ship Type Classification; Specific Aircraft Type Counting; Specific Vehicle Type Counting; Spatial relation classification; Tree Health Assessment; Trees Counting; Water Bodies Counting.\\
\addlinespace
OmniEarth & 19 & 5,656 & \textit{Single-/multi-select MCQ.} Scene Classification; Land-cover Classification; Image Modality Recognition; Object Counting; Fine-grained Category Classification; Attribute Recognition; Spatial Relationship Reasoning; Geometric Measurement; Damage Assessment Reasoning; Seasonal Temporal Reasoning; Geo-localization \& Hydrological Reasoning; Disaster Cause Inference; Geo-Entity Understanding; Planning Suggestions; City Recognition; Image Condition Assessment; Degraded-condition VQA; Hallucination Detection; Semantic consistency.\\
\addlinespace
HRVQA & 10 & 2,000 & \textit{Open-ended QA.} Number; Color; Location; Shape; Yes/No; Areas; Size; Transportation; Scene; Sports. Each group contains 200 questions; Shape is restricted to target-present cases.\\
\addlinespace
LHRS-Bench & 11 & 690 & \textit{Single-choice MCQ.} Identity; Color; Orientation; Shape; Area; Resolution; Modality; Location; Distance; Quantity; Reasoning. Original option order, one pass per question; ability tags overlap.\\
\bottomrule
\end{tabular*}
\end{table*}

\noindent\textbf{Multiple choice.}
The CHOICE, GEOBench-VLM, and OmniEarth manifests contain 18,274 records. We preserve original evaluation imagery and group question variants by their base identifiers. Prompts request an option letter for single choice or comma-separated letters for multiple selection. The parser removes thinking blocks, extracts an answer block when present, and normalizes the resulting letters or exact option text. Single-choice predictions are correct if the parsed option matches the reference. Given a predicted set $P$ and the reference set $G$, OmniEarth uses
\begin{equation}
 s(P,G)=
 \begin{cases}
 1, & P=G,\\
 0.5, & \varnothing\ne P\subsetneq G,\\
 0, & \text{otherwise}.
 \end{cases}
\end{equation}
We evaluate all 690 LHRS-Bench questions once in their original option order. Ability tags can overlap, and overall accuracy counts each question once.

\noindent\textbf{Open-ended answers.}
On the fixed HRVQA subset, we normalize answer case, punctuation, spacing, and number expressions before applying a rule-based answer equivalence rubric.

\subsection{Image Deduplication Checks}
\label{app:dedup}
We compared 5,394 training image files against 19,428 distinct image paths from the five evaluation manifests. SHA-256 hashes of decoded RGB pixels and dimensions found no exact matches. A 64-bit perceptual hash flagged three pairs at Hamming distance at most five. Visual inspection found different scenes in all three pairs, whose CLIP ViT-L/14 cosine similarities ranged from 0.54 to 0.65.

\section{Hyperparameter Sensitivity and Ablations}
\label{app:baselines}

\subsection{Visual Contrast Coefficient}
Table~\ref{tab:lambda} reports the recorded 4B coefficient sweep.

\begin{table}[t]
\centering\small
\captionsetup{skip=4pt}
\caption{\textbf{Sensitivity to the visual contrast coefficient $\lambda$ (4B).} Scores are percentages. \textit{Avg.} weights the five benchmarks equally.}
\label{tab:lambda}
\setlength{\tabcolsep}{4pt}
\resizebox{\columnwidth}{!}{%
\begin{tabular}{@{}lrrrrrr@{}}
\toprule
$\lambda$ & CHOICE & GEO & Omni & HRVQA & LHRS & Avg.\\
\midrule
0.00 & 81.49 & 52.73 & 54.07 & 48.50 & 79.13 & 63.18\\
0.01 & 81.25 & 50.79 & 54.98 & 49.70 & 78.41 & 63.02\\
0.02 & 81.39 & 52.85 & 55.84 & 49.20 & 79.71 & \textbf{63.80}\\
0.04 & 80.58 & 51.17 & 54.92 & 49.95 & 78.12 & 62.95\\
\bottomrule
\end{tabular}%
}
\end{table}

\subsection{Baseline Controls}

\noindent\textbf{Baseline inputs and fact-augmented SFT.}
OPSD~\cite{agarwal2024policy} uses the same image and question for both models and matches their token distributions with unweighted JS. The \origans{} control also supplies the reference answer to the teacher. To test visual facts as direct supervision, we train a 4B SFT baseline to generate the verified fact before the answer on the same 56-step schedule. Its five-benchmark average is lower than answer-only SFT (Table~\ref{tab:sft-facts}).

\begin{table}[t]
\centering\small
\captionsetup{skip=4pt}
\caption{\textbf{Answer-only SFT, fact-augmented SFT, and GeOPD at 4B.} Scores are percentages.}
\label{tab:sft-facts}
\setlength{\tabcolsep}{3.2pt}
\resizebox{\columnwidth}{!}{%
\begin{tabular}{@{}lrrrrrr@{}}
\toprule
Method & CHOICE & GEO & Omni & HRVQA & LHRS & Avg.\\
\midrule
SFT (Answer only) & 81.15 & 50.36 & 53.85 & 46.95 & 78.26 & 62.12\\
SFT (Facts + Answer) & 79.64 & 48.12 & 51.20 & 44.95 & 79.42 & 60.67\\
\textbf{GeOPD (Ours)} & \textbf{81.39} & \textbf{52.85} & \textbf{55.84} & \textbf{49.20} & \textbf{79.71} & \textbf{63.80}\\
\bottomrule
\end{tabular}%
}
\end{table}

\noindent\textbf{Reference-preserving and joint masking.}
On 1,626 training questions with eligible target masks and explicit reference boxes, \textit{Marked + facts (w/o ref.)} masks both regions in the student's masked view, with the same teacher inputs and optimization settings (Table~\docref{tab:ablation}{3}). The joint mask adds a median of 2.35\% of the image area. Relative to joint masking, retaining references raises the five-benchmark average from 63.47\% to 63.80\%, improving object counting (+4.20 on CHOICE, +4.33 on OmniEarth) and spatial relations (+3.67 on CHOICE, +4.13 on OmniEarth), while the aggregate GEOBench-VLM score is 1.62 points lower.

\noindent\textbf{Baseline adaptations.}
Vision-OPD uses a support crop and an EMA teacher. RP-OPSD halves the student's image dimensions while preserving the teacher view and uses bias-corrected reverse KL over the teacher's top 100 tokens. Their EMA update rate is 0.05. RS-OPSD combines global, expanded-context, and detailed views with a correctness-aligned objective, a teacher reliability probe, and a fixed KL reference. Short-answer reliability uses conservative greedy agreement with the canonical answer. GRPO uses the shared answer-equivalence rules and a reference KL coefficient of 0.001. VAD uses a fixed initial teacher, a positive budget of 0.8, and a nonvisual anchor of 0.1. The negative view reduces the support crop to 10\% of its size before resizing it back, with the full search area preserved for global, absence, and counting questions. EAD and S-OPD follow their published visual-alignment and cross-view contrast objectives on the same annotations. SFT uses FP32 parameters, BF16 computation, and one normalization over the accumulated batch of reference answers.

\subsection{One-Sided Visual Gain}
\label{app:contrast}
The detached teacher score drop weights the comparison of the same sampled token across views~\cite{liu2026better}. The one-sided gain in Section~\docref{sec:contrast}{4.3} uses $c=20$ and $K=10$. The corresponding loss has a nonpositive derivative with respect to the original-view score. The derivative is zero when $\ell_{in}^{\mathrm{S}}\le\ell_{in}^{\mathrm{M}}$ or the gain saturates.

\section{Task and Evidence Diagnostics}
\label{app:analysis}

Figure~\docref{fig:training-transfer}{5}(c) compares GeOPD with \origans{} on selected subtasks. Its GEOBench-VLM counting group pools eight native tasks.

\subsection{Evidence Masking Controls}
To construct the held-out masking controls for Figure~\docref{fig:training-transfer}{5}(a), we evaluate 150 questions covering attributes, locations, and spatial relations built from OpenEarthMap~\cite{xia2023openearthmap} and RefSegRS~\cite{yuan2024rrsis} after near-duplicate screening against the training imagery. For each question, GeOPD-4B scores the reference answer on the original view, a target-masked view, and a shape- and area-matched unrelated control view (mean-RGB fill, with spatial references visible), averaging log probability over the answer tokens. Across all 150 questions, target masking lowers the reference-answer score by $0.496\,\text{nats/token}$, compared to $-0.020\,\text{nats/token}$ under unrelated control masking. On relational items with disjoint target and reference masks, occluding only the spatial reference region (while keeping the target visible) lowers the score by $0.793\,\text{nats/token}$, rising to $0.869\,\text{nats/token}$ under joint target-plus-reference masking, confirming reliance on both scene elements.

\subsection{Complete Task Results}
Tables~\ref{tab:alltasks-choice}--\ref{tab:alltasks-lhrs} report per-task scores (\%) on the same records for all three Qwen3.5-4B checkpoints. \textit{Orig+Ans} denotes \origans{}. For LHRS-Bench, $N$ counts questions carrying each tag.

\begin{table*}[!tp]
\centering\small
\captionsetup{skip=4pt}
\caption{\textbf{CHOICE.} Option accuracy across 21 tasks.}
\label{tab:alltasks-choice}
\setlength{\tabcolsep}{6pt}
\renewcommand{\arraystretch}{1.13}
\begin{tabular}{@{}lrrrr@{}}\toprule
Task & $N$ & Base & \textit{Orig+Ans} & GeOPD\\\midrule
attribute comparison & 350 & 88.29 & 87.14 & 87.43\\
attribute recognition & 500 & 60.20 & 65.80 & 64.80\\
change detection & 200 & 72.00 & 71.50 & 69.50\\
disaster discrimination & 200 & 84.50 & 90.50 & 92.00\\
environmental assessment & 200 & 46.50 & 44.00 & 47.00\\
geospatial determination & 200 & 86.50 & 92.00 & 88.50\\
hallucination detection & 500 & 88.80 & 90.00 & 91.00\\
image caption & 200 & 95.50 & 98.00 & 97.50\\
image modality & 800 & 71.75 & 78.00 & 79.12\\
image quality & 800 & 64.12 & 68.75 & 68.62\\
landmark recognition & 100 & 91.00 & 91.00 & 92.00\\
map recognition & 157 & 30.57 & 26.75 & 29.94\\
object counting & 500 & 61.00 & 61.40 & 63.60\\
object localization & 500 & 85.00 & 86.00 & 85.60\\
object presence & 500 & 97.60 & 95.20 & 95.40\\
physical property & 300 & 74.67 & 70.67 & 70.00\\
resource assessment & 600 & 66.50 & 72.67 & 77.00\\
scene classification & 2000 & 95.05 & 95.45 & 95.45\\
situation inference & 300 & 86.33 & 90.67 & 87.33\\
spatial relationship & 300 & 72.67 & 87.67 & 87.33\\
time property & 200 & 58.50 & 65.00 & 66.50\\
\bottomrule
\end{tabular}
\end{table*}

\begin{table*}[!tp]
\centering\small
\captionsetup{skip=4pt}
\caption{\textbf{GEOBench-VLM.} Option accuracy across the 17 evaluated single-image tasks.}
\label{tab:alltasks-geo}
\setlength{\tabcolsep}{6pt}
\renewcommand{\arraystretch}{1.13}
\begin{tabular}{@{}lrrrr@{}}\toprule
Task & $N$ & Base & \textit{Orig+Ans} & GeOPD\\\midrule
Aircraft Type Classification & 100 & 51.00 & 54.00 & 62.00\\
Building Counting & 170 & 26.47 & 34.12 & 47.06\\
Crop Type Classification & 55 & 18.18 & 25.45 & 23.64\\
Disaster Type Classification & 227 & 56.39 & 58.15 & 55.95\\
Fire Risk Assessment & 120 & 17.50 & 14.17 & 13.33\\
General Aircraft Counting & 100 & 31.00 & 38.00 & 45.00\\
General Vehicle Counting & 150 & 22.67 & 29.33 & 46.67\\
Land Use Classification & 410 & 60.24 & 55.12 & 52.68\\
Marine Debris Counting & 100 & 36.00 & 34.00 & 29.00\\
Scene Classification & 415 & 80.96 & 81.20 & 80.00\\
Ship Type Classification & 207 & 40.10 & 51.21 & 48.31\\
Specific Aircraft Type Counting & 140 & 34.29 & 29.29 & 35.00\\
Specific Vehicle Type Counting & 224 & 18.75 & 18.75 & 33.93\\
Spatial Relation Classification & 466 & 75.75 & 79.61 & 75.11\\
Tree Health Assessment & 157 & 24.20 & 15.92 & 33.76\\
Trees Counting & 85 & 22.35 & 17.65 & 28.24\\
Water Bodies Counting & 85 & 56.47 & 61.18 & 64.71\\
\bottomrule
\end{tabular}
\end{table*}

\begin{table*}[!tp]
\centering\small
\captionsetup{skip=4pt}
\caption{\textbf{OmniEarth.} Scores across 19 tasks in the original option order. Multiple selection questions use partial credit.}
\label{tab:alltasks-omni}
\setlength{\tabcolsep}{6pt}
\renewcommand{\arraystretch}{1.13}
\begin{tabular}{@{}lrrrr@{}}\toprule
Task & $N$ & Base & \textit{Orig+Ans} & GeOPD\\\midrule
A1.1-Scene Classification & 400 & 57.25 & 58.50 & 59.25\\
A1.2-Land-cover Classification & 200 & 78.50 & 85.50 & 82.50\\
A1.3-Image Modality Recognition & 140 & 66.43 & 81.43 & 77.86\\
A2.3-Object Counting & 600 & 49.83 & 53.17 & 51.33\\
A2.4-Fine-grained Category Classification & 365 & 45.48 & 51.51 & 51.78\\
A2.5-Attribute Recognition & 300 & 50.67 & 61.67 & 61.67\\
B1.1-Spatial Relationship Reasoning & 315 & 86.67 & 92.38 & 92.06\\
B1.2-Geometric Measurement & 300 & 14.33 & 9.33 & 27.00\\
B2.2-Damage Assessment Reasoning & 381 & 30.71 & 34.65 & 39.63\\
B2.4-Seasonal Temporal Reasoning & 101 & 36.63 & 37.62 & 41.58\\
B3.1-Geo-localization \& Hydrological Reasoning & 237 & 41.77 & 47.26 & 47.26\\
B3.2-Disaster Cause Inference & 525 & 22.67 & 33.90 & 33.71\\
B3.3-Geo-Entity Understanding & 150 & 76.00 & 79.33 & 79.33\\
B3.4-Planning Suggestions & 220 & 30.00 & 38.18 & 37.73\\
B3.5 City Recognition & 322 & 56.83 & 63.98 & 63.35\\
C1.1-Image Condition Assessment & 200 & 55.00 & 66.50 & 72.00\\
C1.2-Degraded-condition VQA & 500 & 53.80 & 64.40 & 61.80\\
C2.1-Hallucination Detection & 100 & 44.50 & 55.00 & 55.50\\
C2.2-Semantic consistency & 300 & 57.00 & 64.00 & 66.00\\
\bottomrule
\end{tabular}
\end{table*}

\begin{table*}[!tp]
\setlength{\tabcolsep}{4pt}
\renewcommand{\arraystretch}{1.13}
\begin{minipage}[t]{\columnwidth}
\vspace{0pt}
\centering\small
\captionsetup{skip=4pt}
\caption{\textbf{HRVQA.} Open-ended answer accuracy across the ten categories of the balanced diagnostic subset.}
\label{tab:alltasks-hrvqa}
\begin{tabular}{@{}lrrrr@{}}\toprule
Task & $N$ & Base & \textit{Orig+Ans} & GeOPD\\\midrule
Number & 200 & 69.00 & 73.00 & 74.50\\
Color & 200 & 11.50 & 27.50 & 37.00\\
Location & 200 & 45.50 & 54.00 & 59.00\\
Shape & 200 & 21.50 & 20.00 & 19.00\\
Yes/No & 200 & 73.00 & 85.00 & 87.00\\
Areas & 200 & 44.00 & 46.50 & 55.50\\
Size & 200 & 40.00 & 38.00 & 64.00\\
Transportation & 200 & 62.00 & 73.00 & 67.50\\
Scene & 200 & 2.50 & 15.00 & 18.00\\
Sports & 200 & 16.50 & 9.50 & 10.50\\
\bottomrule
\end{tabular}
\end{minipage}\hfill
\begin{minipage}[t]{\columnwidth}
\vspace{0pt}
\centering\small
\captionsetup{skip=4pt}
\caption{\textbf{LHRS-Bench.} Option accuracy for all 11 ability tags. The same question may appear under multiple tags.}
\label{tab:alltasks-lhrs}
\begin{tabular}{@{}lrrrr@{}}\toprule
Task & $N$ & Base & \textit{Orig+Ans} & GeOPD\\\midrule
Identity & 634 & 76.66 & 77.60 & 79.18\\
Color & 113 & 80.53 & 79.65 & 84.07\\
Orientation & 39 & 35.90 & 56.41 & 53.85\\
Shape & 37 & 81.08 & 78.38 & 86.49\\
Area & 75 & 73.33 & 82.67 & 81.33\\
Resolution & 21 & 57.14 & 57.14 & 80.95\\
Modality & 23 & 56.52 & 100.00 & 91.30\\
Location & 204 & 82.35 & 84.80 & 84.31\\
Distance & 22 & 59.09 & 54.55 & 63.64\\
Quantity & 137 & 56.20 & 54.01 & 58.39\\
Reasoning & 46 & 80.43 & 76.09 & 76.09\\
\bottomrule
\end{tabular}
\end{minipage}
\end{table*}

\end{document}